\documentclass[letterpaper, 10 pt, conference]{IEEEtran}
\IEEEoverridecommandlockouts

\usepackage{cite}
\usepackage{amsmath,amssymb,amsfonts}
\usepackage{algorithm}
\usepackage{algpseudocode}
\usepackage{graphicx}
\usepackage{textcomp}
\usepackage{xcolor}

\ifCLASSOPTIONcompsoc
\usepackage[caption=false,font=normalsize,labelfont=sf,textfont=sf]{subfig}
\else
\usepackage[caption=false,font=footnotesize]{subfig}
\fi

\def\BibTeX{{\rm B\kern-.05em{\sc i\kern-.025em b}\kern-.08em
    T\kern-.1667em\lower.7ex\hbox{E}\kern-.125emX}}
\begin{document}

\bstctlcite{IEEEexample:BSTcontrol}

\title{
    VFILC: Accurate Frequency Extrapolations in Imitation Learning via Sampling Frequency ILC
\thanks{This work was supported by JSPS KAKENHI Grant Number 24K00905, JST SPRING Grant Number JPMJSP2124, JST PRESTO Grant Number JPMJPR24T3, and JST ALCA-Next Japan Grant Number JPMJAN24F1. This study was based on the results obtained from the JPNP20004 project subsidized by the New Energy and Industrial Technology Development Organization (NEDO).}
}

\author{


    \IEEEauthorblockN{Nozomu Masuya}
    \IEEEauthorblockA{\textit{Grad. School of Science and Technology} \\
    \textit{University of Tsukuba}\\
    Tsukuba, Japan \\
    masuya.nozomu.sm@alumni.tsukuba.ac.jp}
    \and
    \IEEEauthorblockN{Toshiaki Tsuji}
    \IEEEauthorblockA{\textit{Grad. School of Science and Engineering} \\
    \textit{Saitama University}\\
    Saitama, Japan \\
    tsuji@ees.saitama-u.ac.jp}
    \and
    \IEEEauthorblockN{Sho Sakaino}
    \IEEEauthorblockA{\textit{Systems and Information Engineering} \\
    \textit{University of Tsukuba}\\
    Tsukuba, Japan \\
    sakaino@iit.tsukuba.ac.jp}
}

\maketitle

\begin{abstract}
    Conventional neural network~(NN)-based imitation learning methods for variable-speed motion either restricted their scope to interpolated speeds, or generated unpredictable motions when extrapolating beyond trained velocity ranges.
    Variable-frequency imitation learning~(VFIL) enabled extrapolations of speeds by linking the NN model's sampling frequency to the motion frequency, whereas its open-loop configuration caused frequency errors, especially in the extrapolated high-frequency settings.
    This study proposes variable-frequency imitation learning with iterative learning control~(VFILC) based on a combination of VFIL and iterative learning control~(ILC) with both feedforward and feedback parts, the former taking advantage of VFIL and the latter adjusting the frequency errors.
    The experimental results showed that the proposed method successfully and accurately extrapolated motion speeds and reduced frequency errors in all three tasks, and that the feedback especially reduced the frequency errors by a remarkable 81\% in the wiping task and 50\% in the shaking task, both compared to simple feedforward VFIL, when extrapolating at double the average speed in the training data. The proposed method also improved accuracy by 27\% compared with VFIL even at an interpolated frequency for a contact-rich mixing task affected by complex friction traits.

\end{abstract}

\begin{IEEEkeywords}
Imitation Learning, Machine Learning for Robot Control, and Force Control.
\end{IEEEkeywords}

\section{Introduction}
    In fields suffering from a shortage of workers or requiring repetitive tasks, artificial intelligence~(AI)-based robot manipulation, namely reinforcement learning and imitation learning, is gaining attention for its higher adaptability and elimination of the need for hard coding.
    Reinforcement learning-based methods~\cite{levine2018, daydreamer} enable learning from a large number of random trials. 
    Despite the low cost of training from simulations, the gap between simulation environments and real environments~\cite{blancomulero2024} obstructs real-world applications of these methods in robot manipulation.
    Imitation learning-based methods~\cite{yang2017, aloha, pi05}, by contrast, learn from human demonstrations. 
    By training machine learning models to generate upcoming command values from past sensory data, trajectories considering hard-to-simulate settings can be generated. However, these methods require large-scale datasets for multi-task problems, including variable-speed motions, which incorporate different upcoming motions for each speed.

    Variable-frequency imitation learning~(VFIL)~\cite{VFIL} reduced this to single-task settings, by first normalizing variable-speed motion to single-speed motion for training neural network~(NN) models, and then changing the sampling frequency of the NN models according to the desired motion frequencies.
    However, this method was verified only on a wiping task, and frequency errors persisted in extrapolated speeds due to the nonlinear nature of contact-rich tasks.

    This paper proposes variable-frequency imitation learning with iterative learning control~(VFILC), VFIL enhanced with iterative learning control~(ILC) to compensate for the frequency errors.
    VFILC distinguishes itself from VFIL in its iterative feedforward and feedback structure, which corrects frequency errors iteratively while retaining the advantage of VFIL.
    The proposed method was experimentally verified on three tasks: wiping, mixing and shaking.
    The experimental results showed that the proposed method successfully extrapolated the frequency range of motion across all three tasks, achieving equally accurate motion frequencies as interpolated ones.
    The feedback from ILC especially reduced frequency errors in settings with high resistance forces with extrapolated frequencies twice as high as the trained motion.

    The key contributions of this paper is summarized as follows.
    \begin{itemize}
        \item We propose VFILC, a novel frequency correction framework for imitation learning, which directly alters the sampling frequency of NN models. 
        \item We proved the adaptability of VFILC in contact-rich tasks at extrapolated high frequencies, reducing the frequency errors even at the twice the frequency of the average training data.
        \item We proved the stability of the base method VFIL against contact-rich tasks other than wiping, with the mixing task that deals with powder's friction traits, and the shaking task that requires accurate picking and firm grasping.
    \end{itemize}

\section{Related Work}

    \subsection{Learning-based Robot Manipulation with Feedback}
        
        For motions requiring a higher autonomy, including trajectory planning, human feedback is useful to prevent failures and to acquire data efficiently. 
        Takahashi~\textit{et al.}'s CHG-DAgger~\cite{chgdagger}, an extension of HG-DAgger by Kelly~\textit{et al.}~\cite{hgdagger}, implemented cooperative control for human interventions and enabled smooth re-training of motion in an imitation learning setting.
        Also in a reinforcement learning-based approach, Chisari~\textit{et al.} implemented human feedback and correction~\cite{correctme}.
        Whereas these methods contribute to a higher quality of overall motion, the reliance on human feedback  can raise concerns about a higher cost of data collection, especially with repetitive tasks with strict frequency commands.

        For simpler tasks whose results can be evaluated quantitatively, feedback can be automated as in other automatic control methods.
        Sato~\textit{et al.} implemented trajectory error feedback into imitation learning models and enabled force control in unlearned position-only trajectories~\cite{sato2025}, by using multi-layer perceptron without recurrent structures. 
        Whereas this method is applicable to correction of force-controlled trajectories, temporal errors from NN models remained untreated in time domain, which interfere with generation of repetitive motion.
        The proposed method of this study, VFILC, distinguishes itself in handling temporal errors in the time domain, by adjusting the sampling frequency of the NN models.

    \subsection{Variable-Speed Motion Generation from Demonstrations}
        Even when handling unknown objects or making contact, humans can adapt to their nonlinear effects and adjust speeds of motion. 
        To achieve this agility on robot systems, a number of methods with different assumptions were introduced.

        Using probabilistic principal component analysis and assuming negligible inertial forces, Perico~\textit{et al.} achieved adjustment of motion speeds~\cite{perico2019, perico2020}.
        Although this method provides relatively explainable motion generation, the assumption behind it limits this method to low-speed settings with small accelerations.

        Also, using force control and simple fast-forwarding of demonstration data, Yokokura~\textit{et al.} achieved variable-speed playback, under the assumption of negligible environmental changes~\cite{yokokura2009}. 

        Using NN models to generate motions, training on variable-speed datasets enables variable-speed outputs, under assumptions that the generated motion is similar to the training data, in terms of both speed and trajectory.
        Saigusa~\textit{et al.} implemented self-supervised learning along with force control~\cite{CRANEX7params} to achieve time-wise accuracy.
        Although this method enabled a higher accuracy in interpolated frequency commands, insufficient data prevented motion generation at extrapolated velocities, even with retraining the NN model with previous motion data.
        By data augmentation, Yamamoto~\textit{et al.} achieved conveyor-picking at variable speeds using downsampled datasets obtained from three demonstrations~\cite{yamamoto2023}, with visual data and a position-based setting.
        Although this method has enabled variable-speed motions including contacts, application of this method to force-controlled settings would require an assumption of negligible nonlinearity.
        
        Masuya~\textit{et al.}'s VFIL relaxed the temporal aspect of this assumption, by normalizing variable-speed motions into single-speed time series for NN models~\cite{VFIL}.
        Although this method enabled motion at extrapolated speeds, its configuration lacked feedback from the generated motion, causing persistent delays in high-frequency settings. 
        Also, this method was only verified on a wiping task, necessitating further verification.
        The proposed method of this paper, VFILC, tackles these delays using feedback control to compensate delays, and the experiment verifies VFIL-based motion in two other task settings, mixing and shaking.

\section{Method: VFILC}
\begin{figure}[tbp]
        \centering
        \resizebox*{8.8cm}{!}{\includegraphics[scale=0.5, bb = 0 0 612 206]{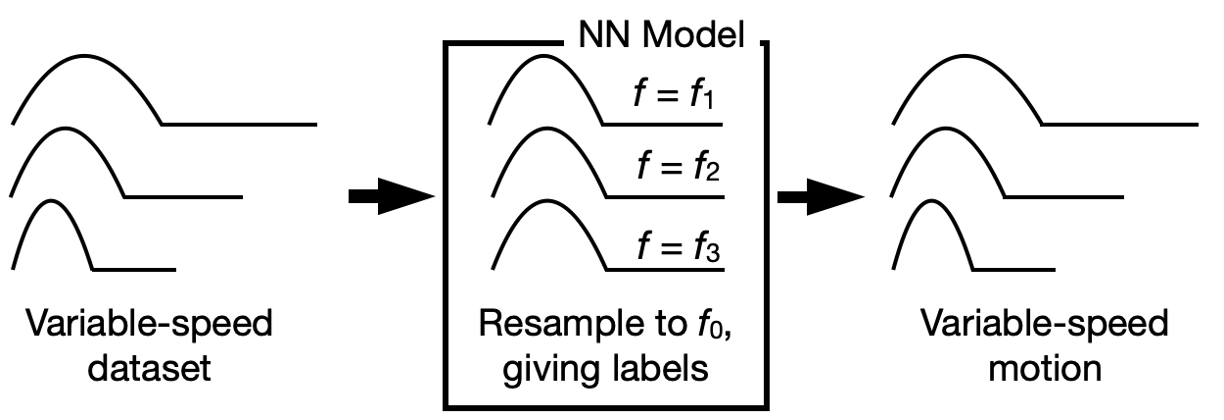}}
        \caption{Concept of plain VFIL.}
        \label{fig:vfil}
    \end{figure}

\begin{figure}[tbp]
        \centering
        \subfloat[Configuration of VFIL]{%
            \resizebox*{8cm}{!}{\includegraphics[scale=0.8, bb = 0 0 612 198]{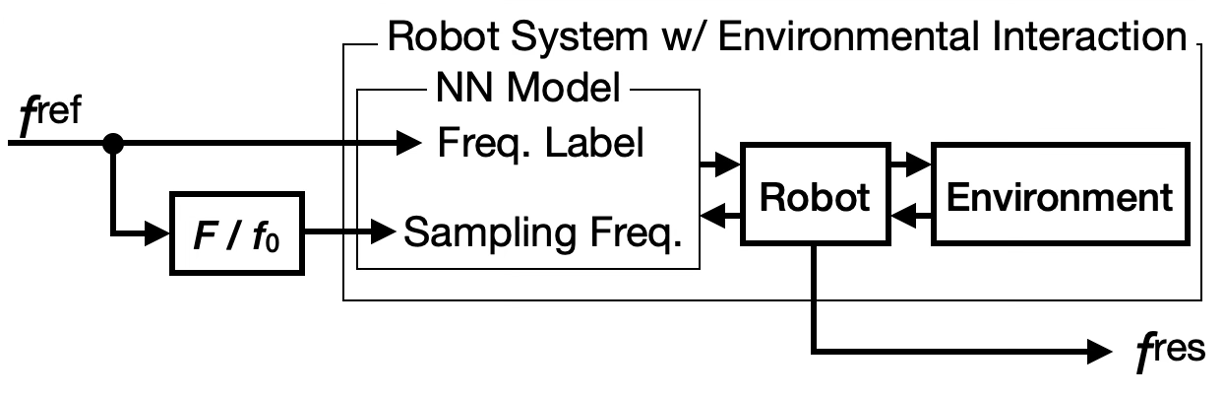}}
            \label{vfil}
        }
        \vspace{0.5pt}
        \subfloat[Configuration of VFILC]{
        \resizebox*{8.8cm}{!}{\includegraphics[scale=0.8, bb = 0 0 612 231]{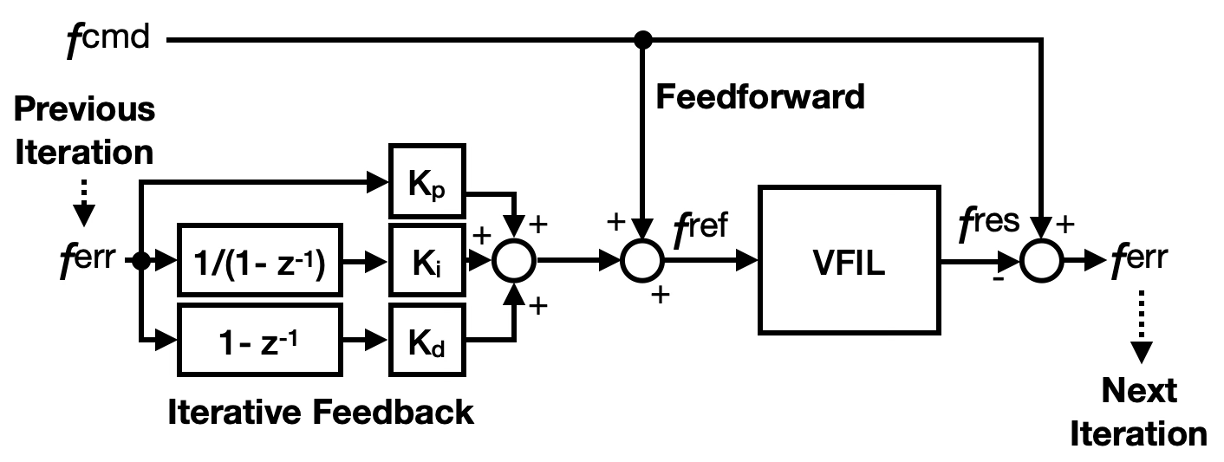}}
            \label{vfilc}
        }
        \caption{Configurations of VFIL and VFILC.}
        \label{fig:vfilc}
    \end{figure}

    This study proposes variable-frequency imitation learning with iterative learning control~(VFILC), an integrated method of VFIL~\cite{VFIL} and iterative learning control~(ILC)~\cite{ILC, ahn2007}, to enable velocity-based feedback and correction.
    In plain VFIL, as shown in Figs.~\ref{fig:vfil} and~\ref{fig:vfilc}\subref{vfil}, the training data were resampled to a single base motion frequency $f_0$ at the corresponding sampling frequency $F$, along with frequency labels indicating the speed-induced changes of environmental conditions.
    In the phase of motion generation at the reference motion frequency $f^{ref}$, the NN model generates variable-speed motions by fast-forwarding the generated motion at $f_0$, with its sampling frequency changing to $f^{ref}F / f_0$. 
    However, this open-loop configuration was vulnerable to nonlinearity-invoked resistance from the environment, which caused the frequency error between $f^{ref}$ and the real motion frequency $f^{res}$, especially when extrapolating frequency ranges.

    To mitigate this error, the proposed method utilizes the frequency error $f^{err}$ from the previous iterations and modifies the VFIL's frequency input $f^{ref}$ using ILC with feedforward and feedback parts, as shown in Fig.~\ref{fig:vfilc}\subref{vfilc}. Variables $K_p$, $K_i$, and $K_d$ correspond to the proportional, integral and derivative control gain each. 
    The feedforward part directly inputs the frequency command $f^{cmd}$ from the first iteration where $f^{err}$ is zero, ensuring $f^{ref} = f^{cmd}$. 
    This feedforward enables a full utilization of VFIL, making the convergence of $f^{res}$ faster and minimizing the risk of excessive feedback inputs, which can lead to drastic changes of $f^{ref}$ and the NN model's sampling frequency.
    In the following iterations, the feedback part based on a discrete-time proportional-integral-derivative~(PID) controller compensates $f^{err}$.

    Therefore, $f^{ref}$ and $f^{err}$ of $i$-th iteration $(i \geq 1)$ , $f^{ref}_i$ and $f^{err}_i$ is derived as
    \begin{equation}
        f^{ref}_i = f^{cmd}_i + K_p f^{err}_{i-1} + K_d (f^{err}_{i-1} - f^{err}_{i-2}) + K_i\sum_{k=0}^{i-1}f^{err}_k
    \end{equation}
    and
    \begin{equation}
        f^{err}_i = f^{cmd}_i  - f^{res}_i 
    \end{equation}
    respectively, with $f^{cmd}$ and $f^{res}$ of $i$-th iteration corresponding to $f^{cmd}_i$ and $f^{res}_i$ each and $f^{err}_{i} = 0 $ for $i \leq 0$.

    To mitigate nonlinearity and uncertainty of NN-generated motions, frequency response $f^{res}$ was calculated as the average motion frequency of each iteration. Therefore, the feedback of the proposed method is conducted solely at the end of each iteration, ensuring that the sampling frequency of NN model does not fluctuate while generating motion.

\section{Experiments and Evaluations}
    To evaluate the feasibility of VFILC, we designed experiments on wiping, mixing and shaking tasks, based on bilateral control-based imitation learning~\cite{CRANEX7params}. 

\begin{figure}[tbp]
        \centering
        \subfloat[Four-channel bilateral control]{%
            \resizebox*{4cm}{!}{\includegraphics[scale=0.35, bb = 0 0 2550 1325]{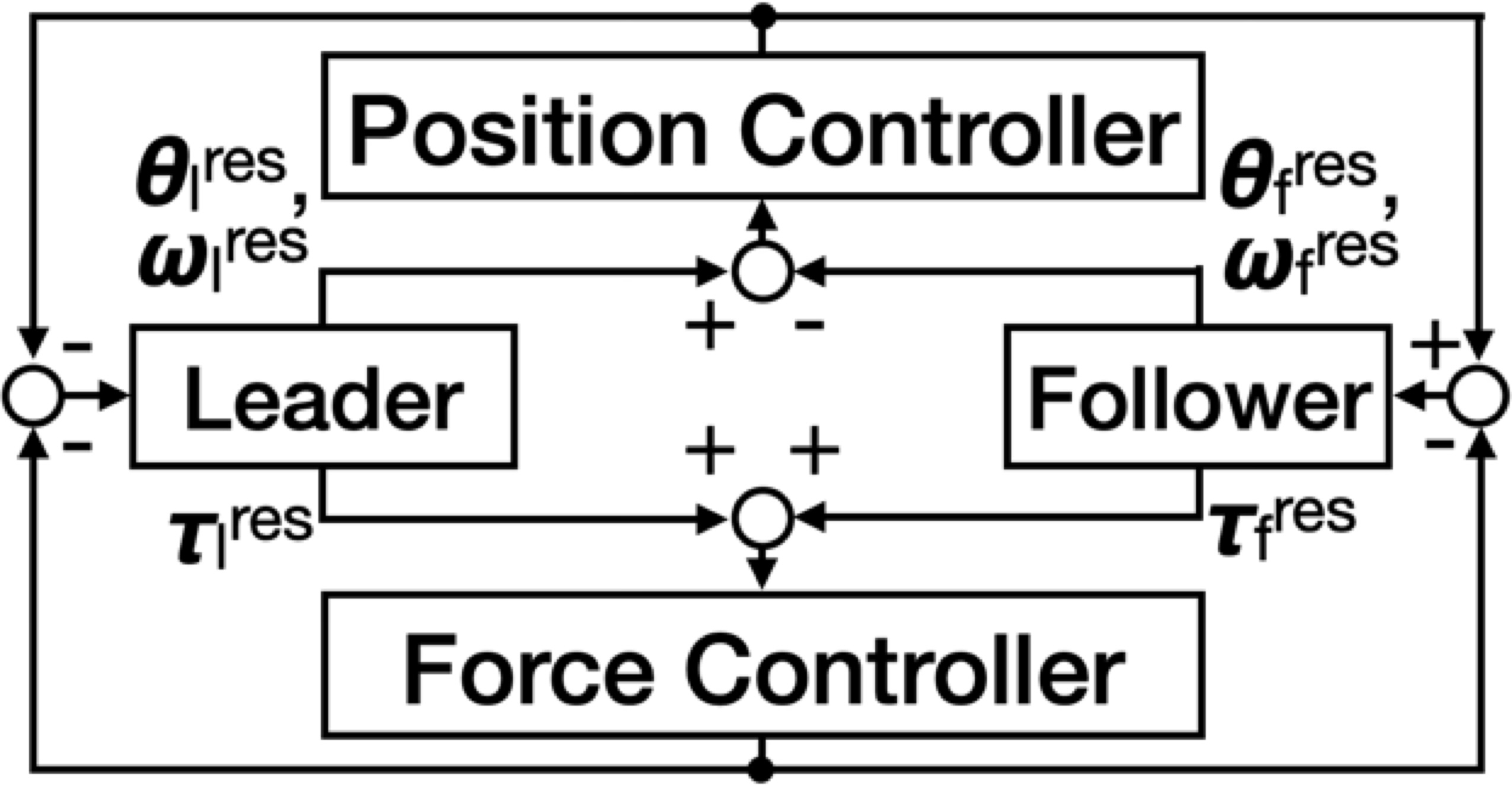}}
            \label{4ch}
        }
        \hspace{1pt}
        \subfloat[Bilateral control-based imitation learning]{%
            \resizebox*{4cm}{!}{\includegraphics[scale=0.35, bb = 0 0 2533 1312]{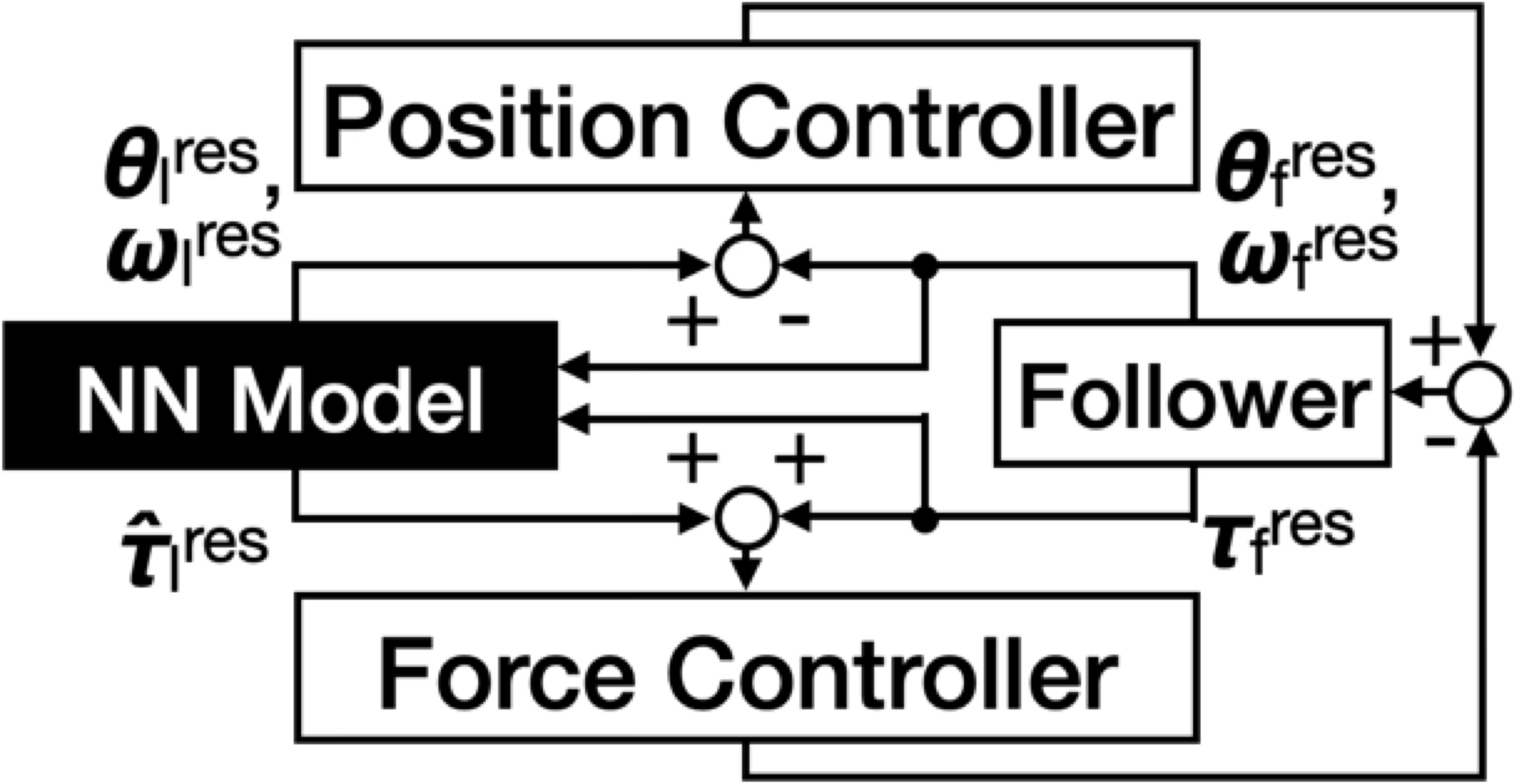}}
            \label{bcil}
        }
        \caption{Block diagrams of four-channel bilateral control and bilateral control-based imitation learning.}
        \label{fig:4chbcil}
    \end{figure}

    In the data collection phase, four-channel bilateral control~\cite{sakaino2011} was used.
    Four-channel bilateral control provides symmetrical position and force feedback, as shown in block diagram Fig.~\ref{fig:4chbcil}\subref{4ch}.
    Variables $\bold{\theta}$, $\bold{\omega}$, and $\bold{\tau}$ signifies joint angle, angular velocity, and torque, respectively.
    Superscript $ref$ indicates the reference value, and $res$ indicates the response value.
    This method virtually replicates the action-reaction law between leader and follower robots, subscripted $l$ and $f$ respectively, by setting
    \begin{equation}
        \bold{\theta}^\mathrm{res}_f - \bold{\theta}^\mathrm{res}_l = \bold {0} 
    \end{equation}
    and
    \begin{equation}
        \bold{\tau}^\mathrm{res}_f + \bold{\tau}^\mathrm{res}_l = \bold {0}
    \end{equation}
    as control objectives.

    After training the NN model using the obtained data, the leader robot was replaced with the NN model, as shown in Fig.~\ref{fig:4chbcil}\subref{bcil}.
    Circumflexes on variables signify the estimated values.

    \subsection{Experimental Setup}
        To evaluate the feasibility of VFILC, wiping, mixing and shaking tasks were prepared. 
        All tasks shared the configurations of robots and learning models, while NN models' training data, weights and ILC gains were different.
        Failed iterations which makes $f^{res}$ inaccurate or incalculable, as defined for each task, were excluded from ILC. 
    
        \subsubsection{Setup of Robots}
\begin{figure}[tbp]
                \centerline{\includegraphics[scale=0.06, bb=0 0 2000 2258]{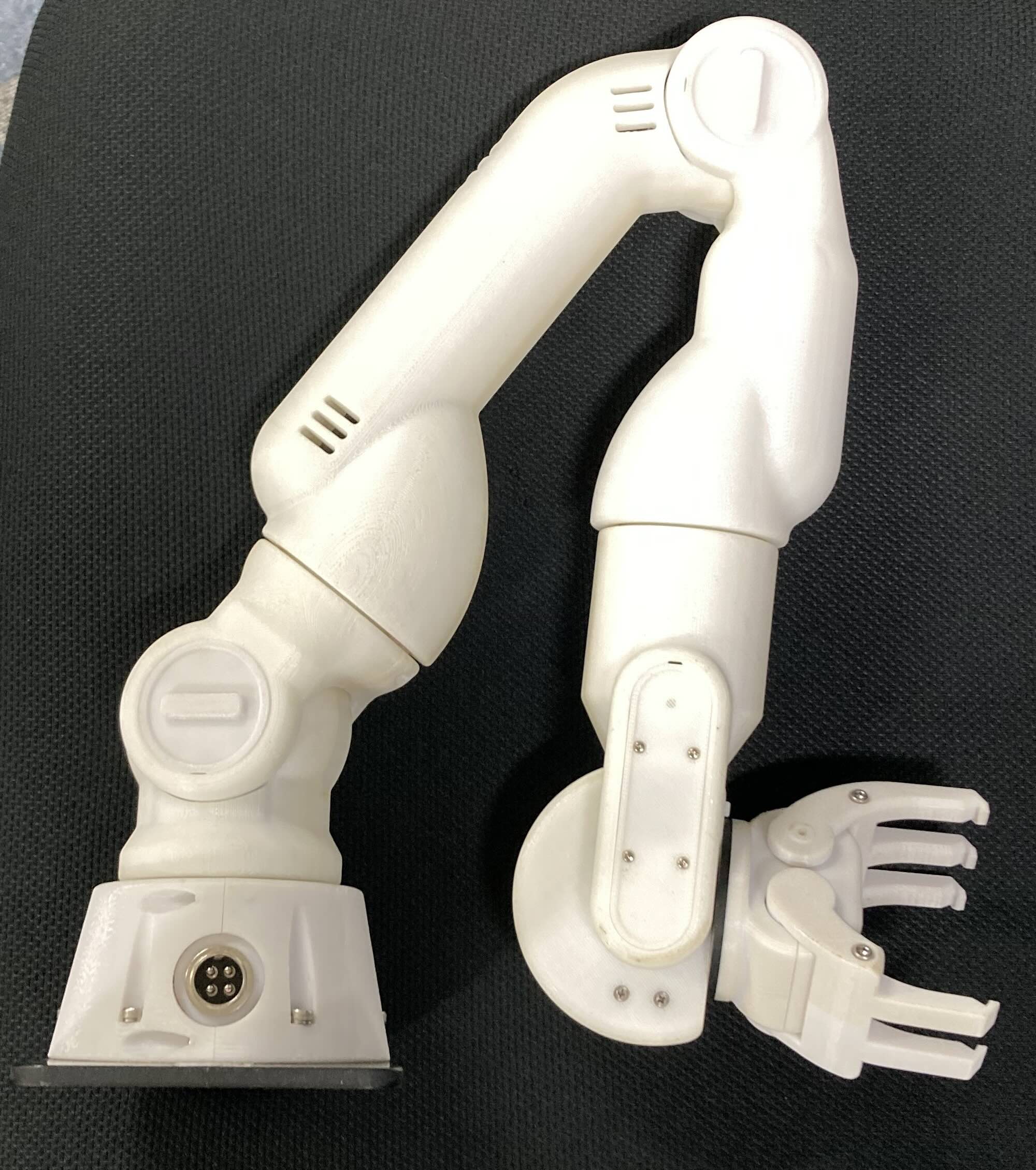}}
                \caption{CRANE-X7 manipulator with a cross-structured hand.}
                \label{fig:crane}
            \end{figure}

            For both the leader and the follower robots, we used two CRANE-X7 manipulators~(RT Corp., Tokyo), as shown in Fig.~\ref{fig:crane}. 
            Each manipulator consisted of a seven-degree-of-freedom~(DoF) arm and a single-DoF rigid end effector.
            The end effectors were replaced with Yamane~\textit{et al.}'s cross-structured hands~\cite{yamaneral} to enable firm grasping, and the third joint of the arms was constrained to eliminate the redundant DoF.
            
\begin{figure}[tbp]
                \centerline{\includegraphics[scale=0.38, bb=0 0 612 222]{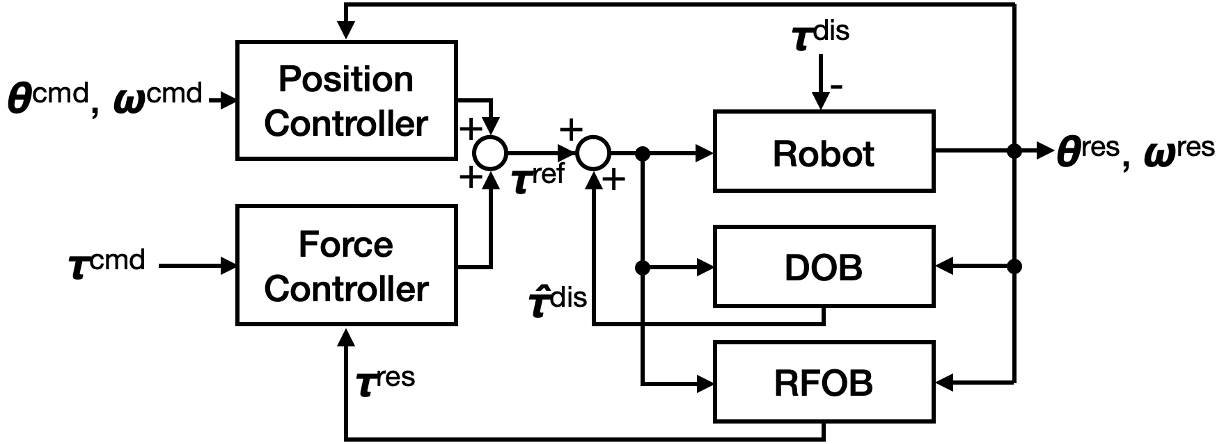}}
                \caption{Block diagram of manipulators' controller.}
                \label{fig:dobrfob}
            \end{figure}

            For sensorless torque sensing and torque control, we utilized a disturbance observer~(DOB)~\cite{DOB} and a reaction force observer~(RFOB)~\cite{RFOB}. 
            Fig.~\ref{fig:dobrfob} describes the controller for each manipulator, equipped with DOB and RFOB.
            Superscripts $cmd$ and $dis$ denote the command and disturbance respectively, and ${\tau^{dis}}$ is the DOB estimate of the disturbance torque.
            Controller parameters were identical to those used by Saigusa \textit{et al.}~\cite{CRANEX7params}.
            
            The sampling frequency of the robot's controller loop and bilateral control was 500~Hz, regardless of the motion frequency of the tasks.
            NN model inputs and outputs, whose frequencies were lower than 100~Hz, were adjusted to the nearest samples and held using zeroth-order hold.

        \subsubsection{Setup of Learning Models}
\begin{figure}[tbp]
                \centerline{\includegraphics[scale=0.4, bb=0 0 612 277]{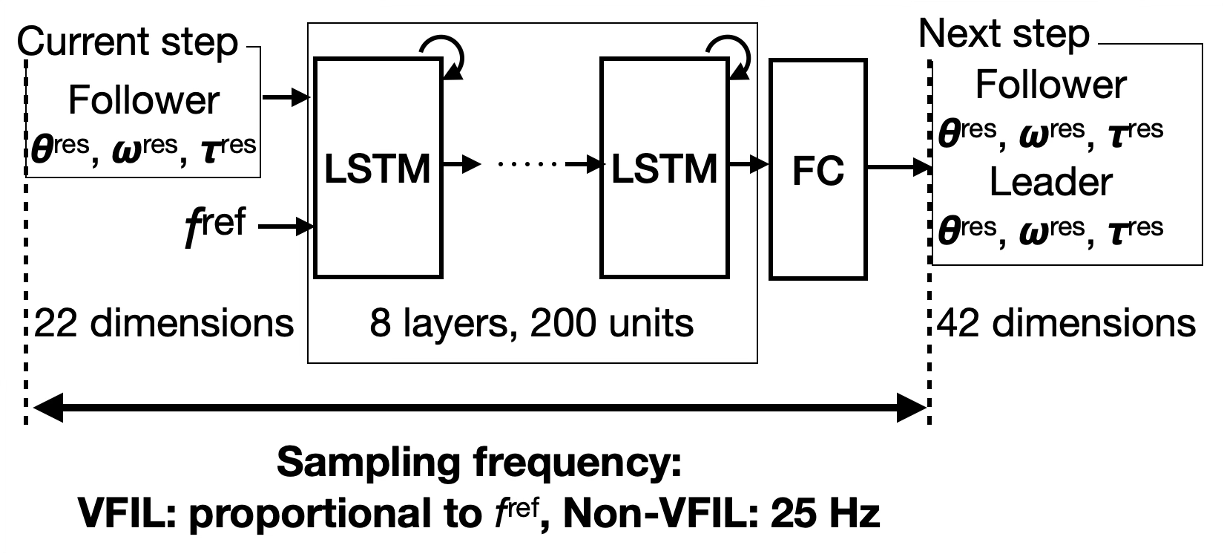}}
                \caption{Configuration of the NN model.}
                \label{fig:nnmodel}
            \end{figure}

            Fig.~\ref{fig:nnmodel} depicts the common configuration of the NN models across all of the three tasks, both the VFIL-based setting and conventional, non-VFIL setting. 
            The models had eight layers, 200 units of long short-term memory~(LSTM) layers~\cite{LSTM1, LSTM2} and one fully connected~(FC) layer.
            The input had 22 dimensions, with one-dimensional frequency command and seven-dimensional follower response values of joint angles, angular velocities and torques each, excluding the constrained joint.
            The output consisted of both the leader and follower responses~\cite{F2FL} of joint angles, angular velocities and torques likewise, to allow for autoregressive validation from predicted follower response values. 
            Superscript $res$ denotes the response value obtained from bilateral control.

            In the VFIL setting, sampling frequencies of the NN models $F$ were 25~Hz at the base motion frequency $f_0$ of each task.
            To resample and generate equal number of episodes from different timings~\cite{rahmatizadeh2018}, each training episode with motion frequency $f$ collected at 500~Hz was first resampled to $500f/f_0$~Hz using linear interpolation, and finally resampled to $25f/f_0$~Hz with 20 different temporal offsets.
            Angular velocities were also scaled to simulate identical trajectories.

            In the non-VFIL setting, NN models' sampling frequency was fixed to 25~Hz, in both training and evaluation phases.
            All training data were resampled to 25~Hz with 20 different temporal offsets, and angular velocities were not adjusted.

            All NN models were trained for 5000 epochs, with the learning rate at $1.0 \times 10^{-4}$ and the batch size at 20.
   
        \subsubsection{Setup of the Wiping Task}
\begin{figure}[tbp]
                \centering
                \subfloat[Initial state]{%
                    \resizebox*{!}{3.1cm}{\includegraphics[scale=0.15, bb = 0 0 534 1080]{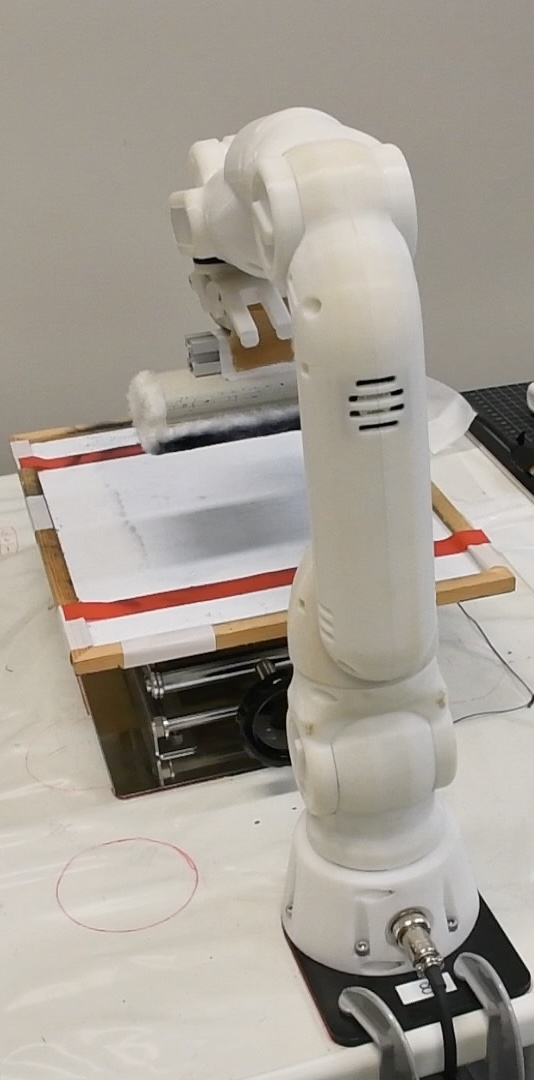}}
                }
                \hspace{5pt}
                \subfloat[Pressing]{%
                    \resizebox*{!}{3.1cm}{\includegraphics[scale=0.15, bb = 0 0 533 1080]{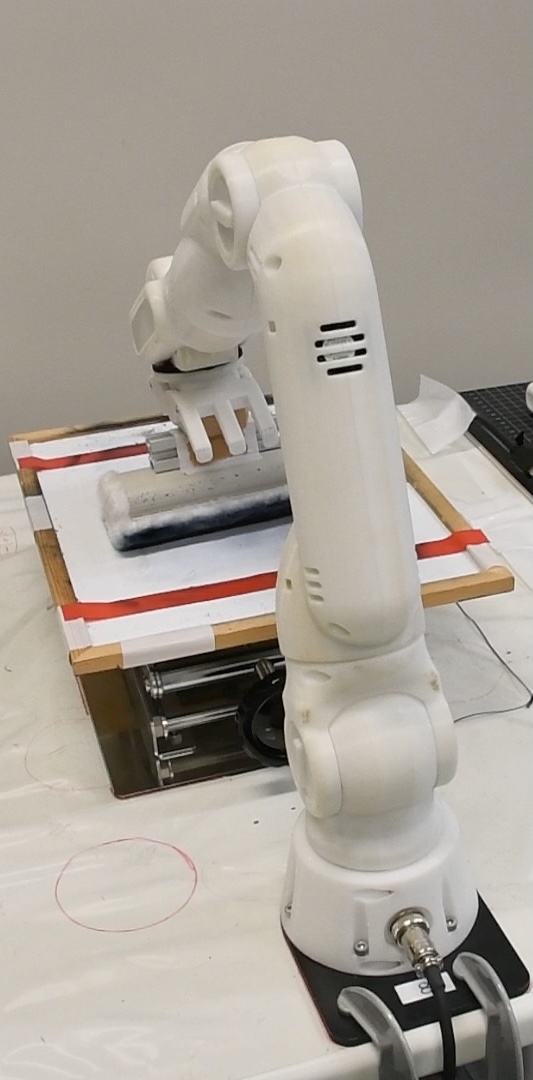}}
                }
                \hspace{5pt}
                \subfloat[Grasping]{%
                    \resizebox*{!}{3.1cm}{\includegraphics[scale=0.15, bb = 0 0 533 1080]{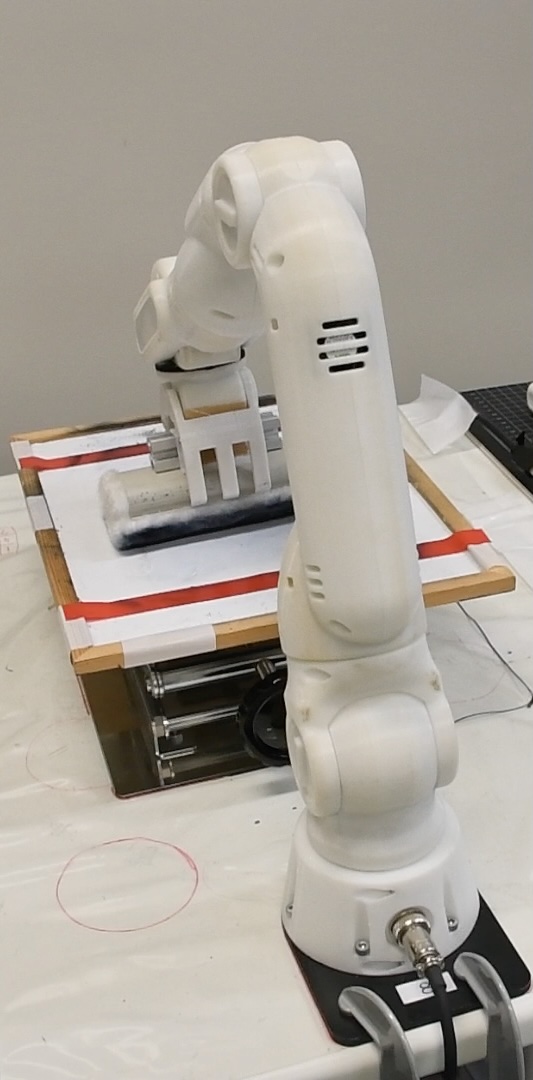}}
                }
                \hspace{5pt}
                \subfloat[Wiping]{%
                    \resizebox*{!}{3.1cm}{\includegraphics[scale=0.15, bb = 0 0 531 1080]{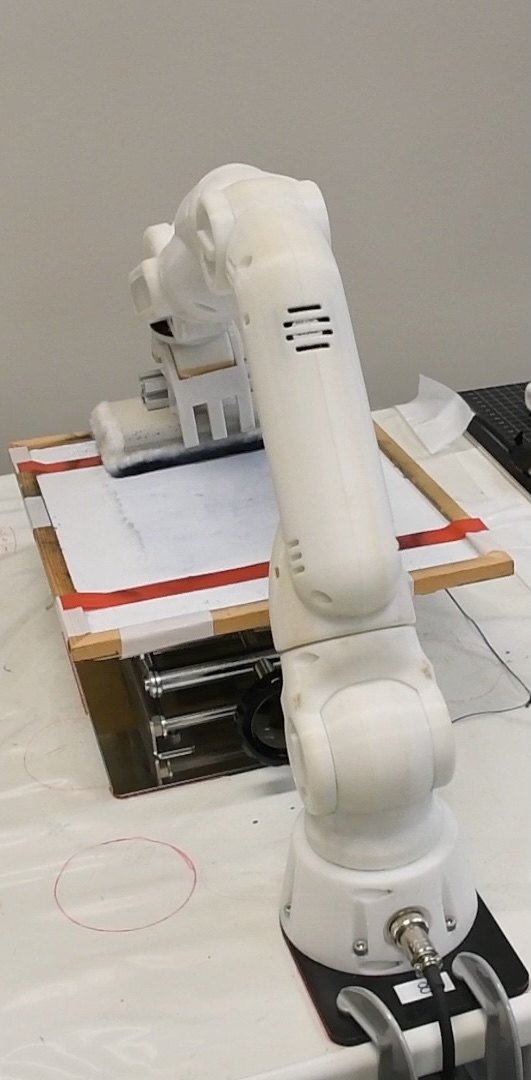}}
                }
                \caption{Procedure of the wiping task.}
                \label{fig:wptask}
            \end{figure}

            To evaluate the fundamental traits of VFILC, the wiping task shown in Fig.~\ref{fig:wptask} was designed.
            The wiping surface was fixed, and an eraser was loosely attached to the gripper.
            First, the robot must first press the eraser to the wiping surface, grasp it, and then continuously wipe the surface at a prescribed frequency.
            An iteration was considered to have failed if the robot did not complete one round trip of wiping within 15~s, or if the robot exceeded the speed limit.

            The teaching process of the task consisted of 18 demonstrations, three demonstrations at each combination of two heights~(10, 15~cm) and three frequencies~(0.4, 0.6, 0.8~Hz), using a metronome. 
            The base motion frequency $f_0$ of this task was 0.6~Hz, tied with NN model's sampling frequency $F=$~ 25~Hz. 
            Data with motion frequencies $f$(= 0.4 and 0.8 Hz) were downsampled to $25f/f_0$(= 16.7 and 33.3~Hz) for the training dataset, respectively.
            
            Trials were conducted at $f=$~0.6 and 1.4~Hz, five times each at 10 and 15~cm height. 
            Each non-VFIL trial consisted of a single 15~s iteration, without ILC.
            Each VFILC trial consisted of 10 iterations of 15~s each, with the first iteration being plain VFIL without feedback.
            In the 1.4~Hz experiment, NN model's initial sampling frequency $fF/f_0$ was 58.3~Hz.
            Real wiping frequency, $f^{res}$, was calculated from the averaged intervals between successive peak values of the second joint angle.
            Four sets of ILC gains were prepared, where the basic set had $K_p$, $K_i$, and $K_d$, set to 0.25, 0.5, and 0.25, respectively derived from the empirical tuning, and the other three sets with doubled $K_p$, $K_i$, or $K_d$ each, i.e., $ \{ K_p, K_i, K_d \} $ =  $\{ 0.5, 0.5, 0.25 \}, \{ 0.25, 1.0, 0.25 \}, \{ 0.25, 0.5, 0.5 \} $, to evaluate the effect of each gain.

        \subsubsection{Setup of the Mixing Task}
\begin{figure}[tbp]
                \centering
                \subfloat[Initial state]{%
                    \resizebox*{!}{3.1cm}{\includegraphics[scale=0.15, bb = 0 0 636 1080]{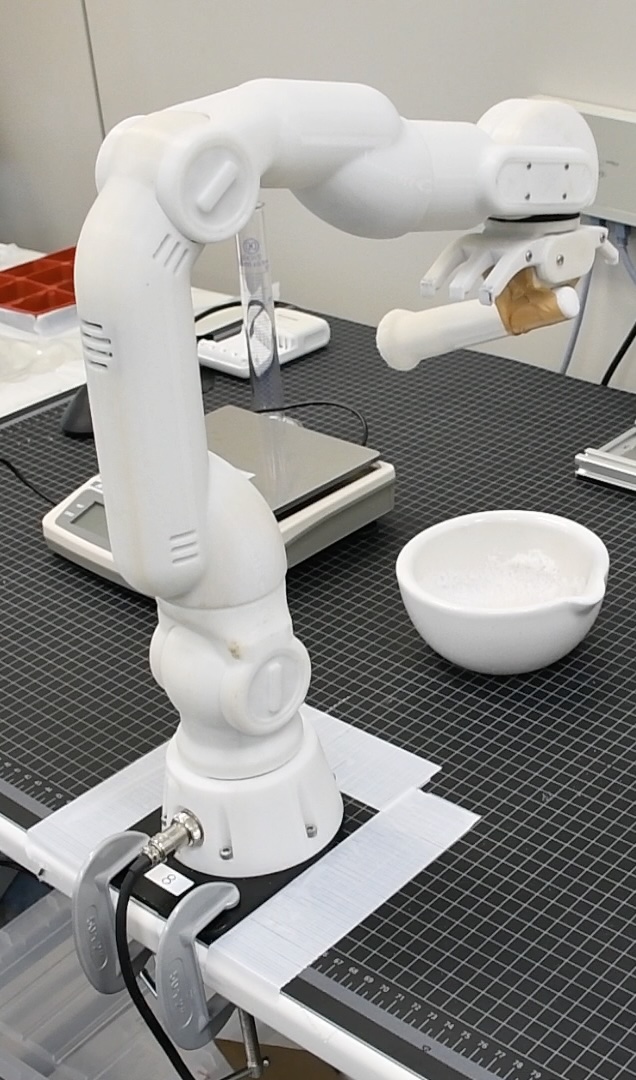}}
                }
                \hspace{5pt}
                \subfloat[Grasping]{%
                    \resizebox*{!}{3.1cm}{\includegraphics[scale=0.15, bb = 0 0 638 1080]{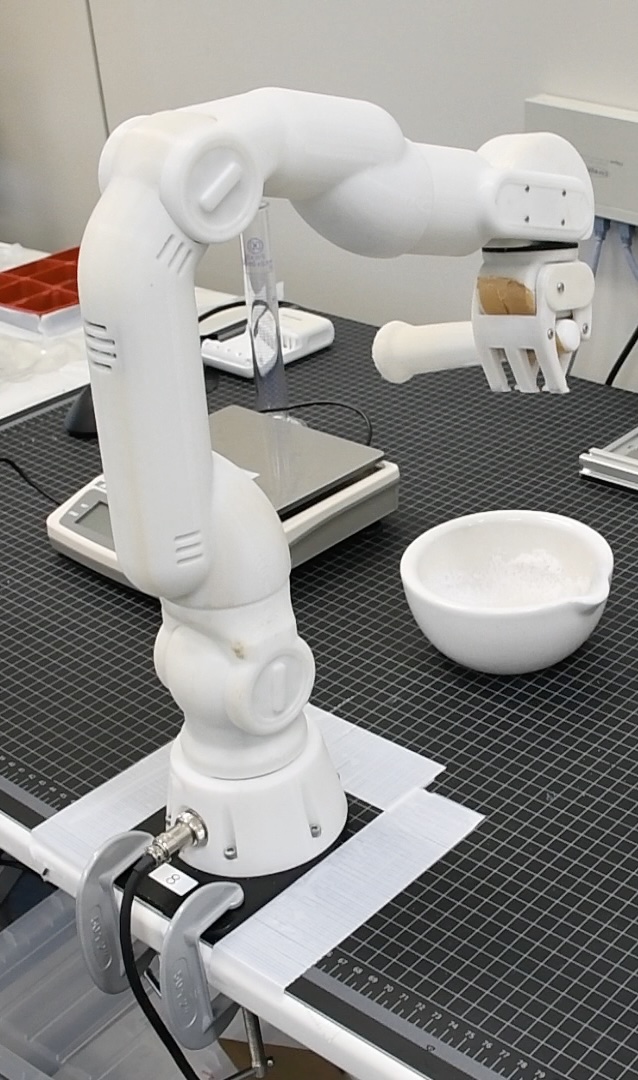}}
                }
                \hspace{5pt}
                \subfloat[Pressing]{%
                    \resizebox*{!}{3.1cm}{\includegraphics[scale=0.15, bb = 0 0 636 1080]{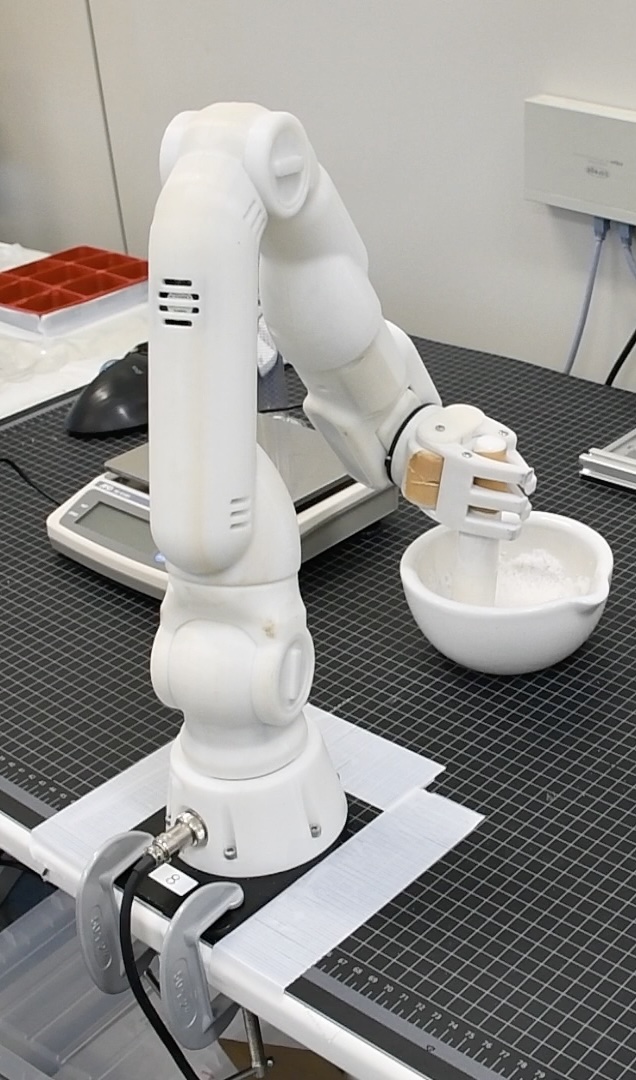}}
                }
                \hspace{5pt}
                \subfloat[Mixing]{%
                    \resizebox*{!}{3.1cm}{\includegraphics[scale=0.15, bb = 0 0 640 1080]{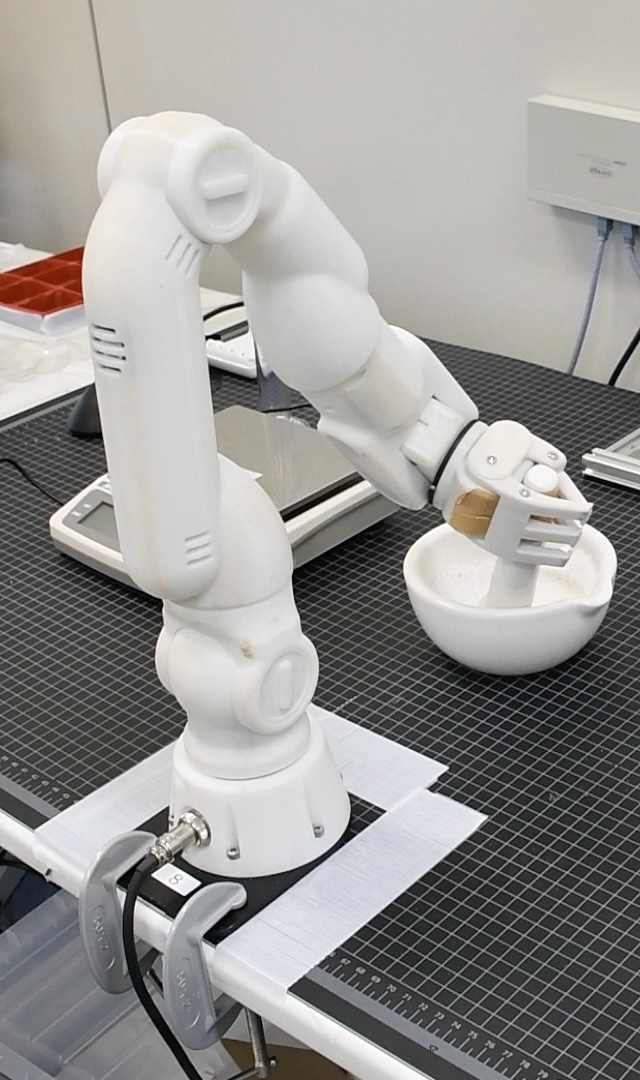}}
                }
                \caption{Procedure of the mixing task.}
                \label{fig:mxtask}
            \end{figure}

            To evaluate under a laboratory-like and contact-rich setting, the mixing task using a pestle and a mortar, shown in Fig.~\ref{fig:mxtask}, was prepared.
            In this task, the robot must first grasp the pestle, place into the mortar, and then continuously mix at a prescribed frequency.
            To enable repetitive experiments without human intervention, the pestle was loosely attached to the robot's gripper.
            An iteration was considered to have failed if the robot did not start mixing within 15~s, or the robot exceeded the speed limit.

            Teaching process of the task consisted of 9 demonstrations with 30~g of salt in the mortar, three times each at 0.4, 0.5, and 0.6~Hz with a metronome. 
            The base motion frequency $f_0$ of this task was 0.5~Hz, tied to NN's sampling frequency $F=$~25~Hz. 
            Data with motion frequencies $f$(= 0.4 and 0.6 Hz) were downsampled to $25f/f_0$(= 20 and 30~Hz) training dataset, respectively.
            
            Trials were conducted at 0.5 and 1.0~Hz, five times each with 30~g of salt only, and 30~g of salt mixed with 50~g of yellow sugar. 
            In the 1.0~Hz experiment, NN model's initial sampling frequency $fF/f_0$ was 50~Hz.
            Each VFILC trial consisted of 10 iterations of 15~s each, whereas each non-VFIL trial consisted of a single 15~s iteration.
            Spilled material was recovered at the end of each trial, or when the total amount of spillage exceeded 20~g. 
            Real mixing frequency, $f^{res}$, was calculated from the durations between the peak angle value of the fourth joint.
            ILC gains $K_p$, $K_i$, and $K_d$ were set to 0.25, 0.5, and 0.25, respectively.
            
        \subsubsection{Setup of the Shaking Task}
\begin{figure}[tbp]
                \centering
                \subfloat[Initial state]{%
                    \resizebox*{!}{2.6cm}{\includegraphics[scale=0.15, bb = 0 0 711 1080]{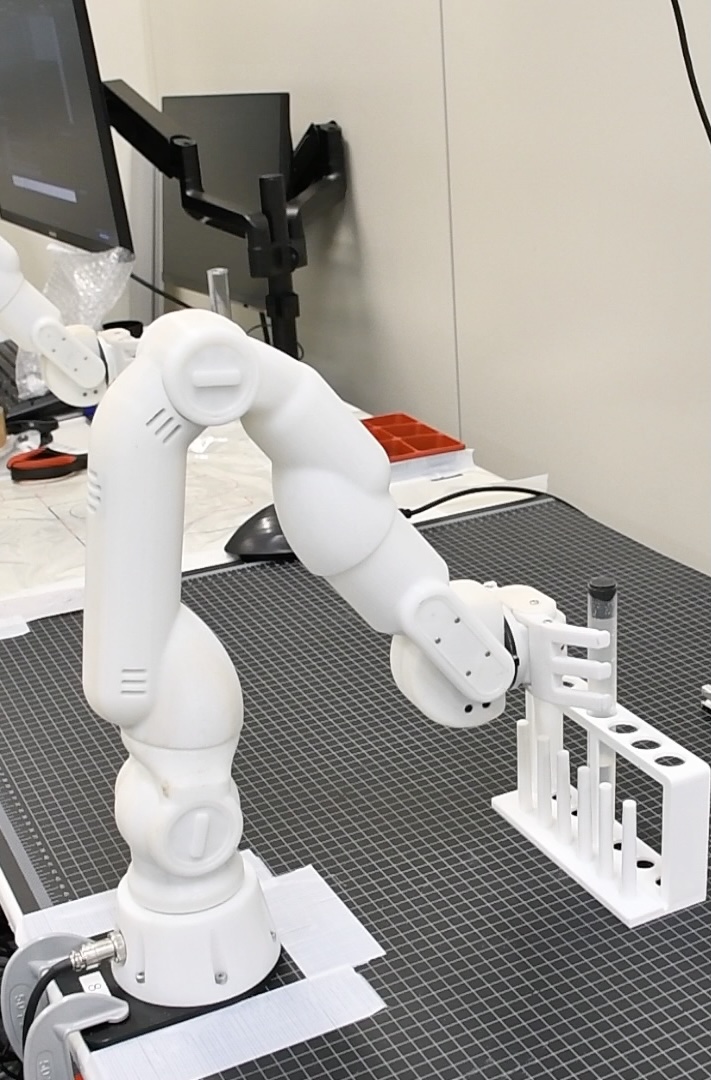}}
                }
                \hspace{5pt}
                \subfloat[Grasping]{%
                    \resizebox*{!}{2.6cm}{\includegraphics[scale=0.15, bb = 0 0 710 1076]{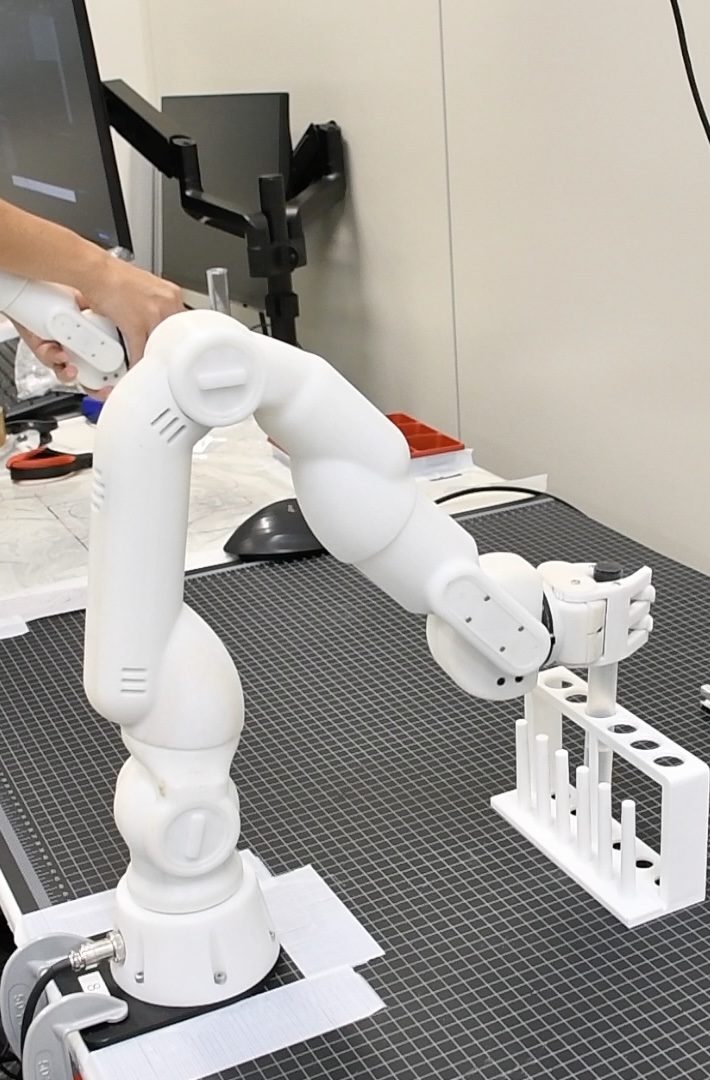}}
                }
                \hspace{5pt}
                \subfloat[Pulling]{%
                    \resizebox*{!}{2.6cm}{\includegraphics[scale=0.15, bb = 0 0 718 1080]{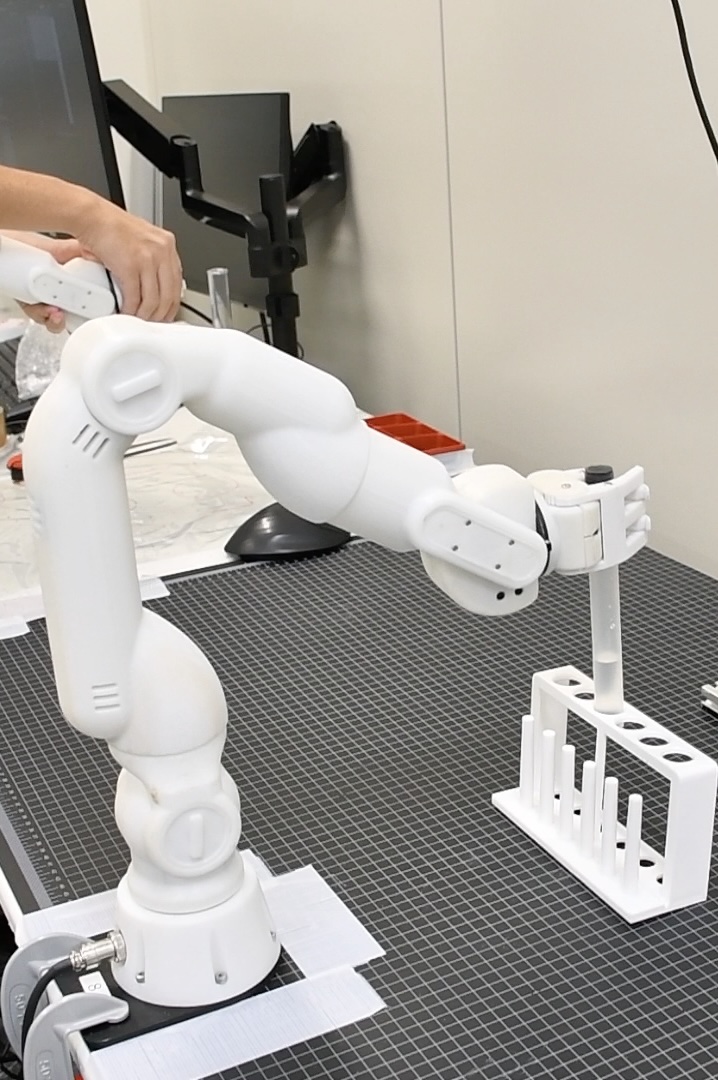}}
                }
                \hspace{5pt}
                \subfloat[Shaking]{%
                    \resizebox*{!}{2.6cm}{\includegraphics[scale=0.15, bb = 0 0 766 1080]{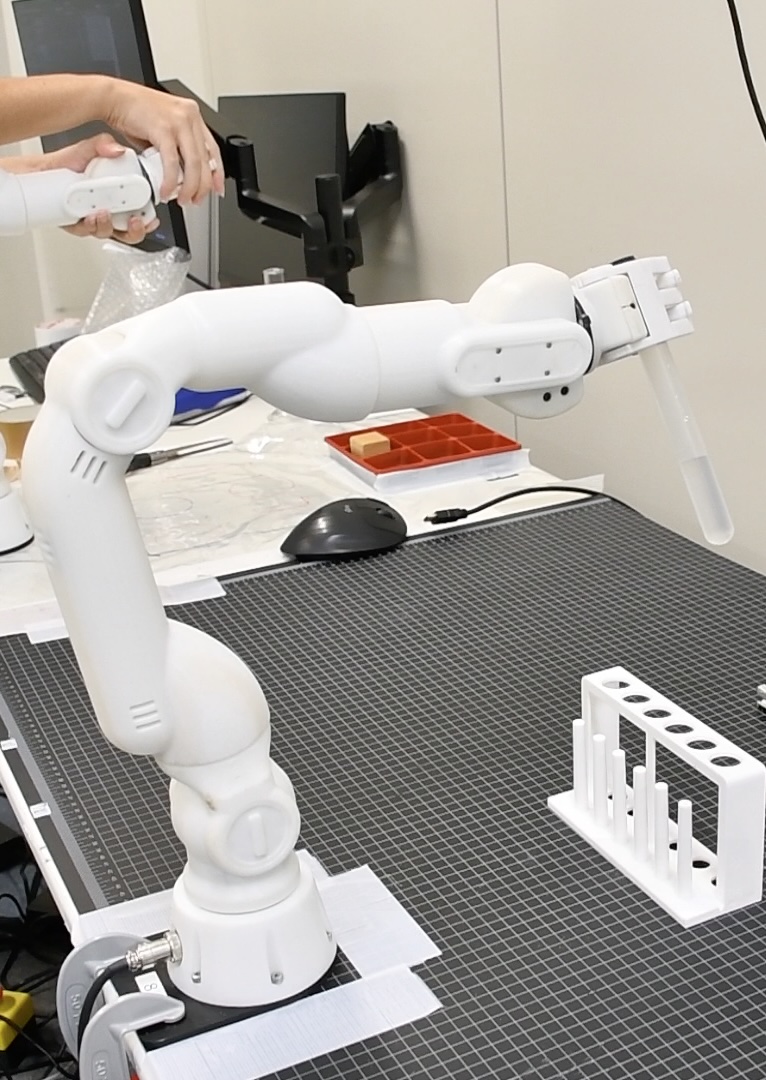}}
                }
                \caption{Procedure of the shaking task.}
                \label{fig:shtask}
            \end{figure}

            To evaluate under a laboratory-like and high-frequency setting, the shaking task using a test tube, shown in Fig.~\ref{fig:shtask}, was prepared.
            In this task, the robot must first grasp the test tube, pull out, and then shake continuously at a prescribed frequency.
            An iteration was considered to have failed if the robot did not grasp the test tube, or did not start shaking within 15~s.

            Teaching process of the task consisted of 9 demonstrations with 10~mL of water in the plugged plastic test tube, three times each at 0.8, 1.0, and 1.2~Hz, using a metronome. 
            To avoid unnecessary stopping, the portions beyond 100~s were removed from training data.
            The base motion frequency $f_0$ of this task, tied to the NN model's $F=$~25~Hz sampling frequency, was 1.0~Hz. 
            Data with motion frequencies $f$(= 0.8 and 1.2 Hz) were downsampled to $25f/f_0$(= 20 and 30~Hz) to form the training dataset, respectively.
            
            Trials were conducted at $f=$~1.0 and 2.0~Hz, five times each at learned 10~mL and unseen 25~mL volume.
            In the 2.0~Hz experiment, NN model's initial sampling frequency $fF/f_0$ was 50~Hz.
            Each VFILC trial consisted of 10 iterations, and each non-VFIL trial consisted of one iteration.
            Each iteration lasted 15~s.
            Real shaking frequency, $f^{res}$, was calculated from the durations between the peak angle values of the seventh joint.
            ILC gains $K_p$, $K_i$, and $K_d$ were set to 0.25, 0.5, and 0.25, respectively.
            
    \subsection{Results}
        \subsubsection{Wiping Task}
\begin{figure}[tbp]
                \centering
                \subfloat[0.6 Hz]{%
                    
                    \resizebox*{8.8cm}{!}{\includegraphics[scale=0.6, bb = 0 0 577 217]{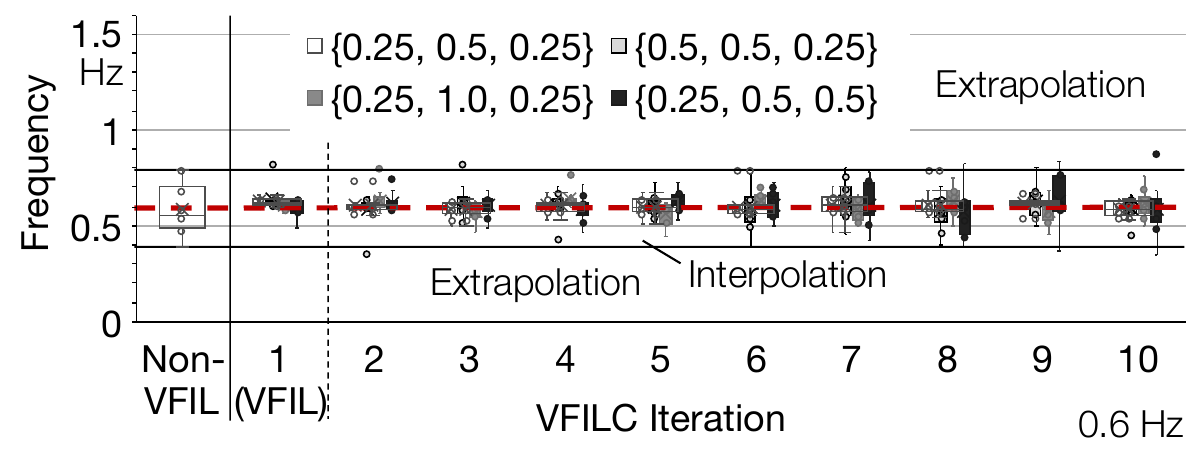}}
                    \label{wp06freq2}
                }
                \vspace{0.5pt}
                \centering
                \subfloat[1.4 Hz]{%
                    
                    \resizebox*{8.8cm}{!}{\includegraphics[scale=0.6, bb = 0 0 577 217]{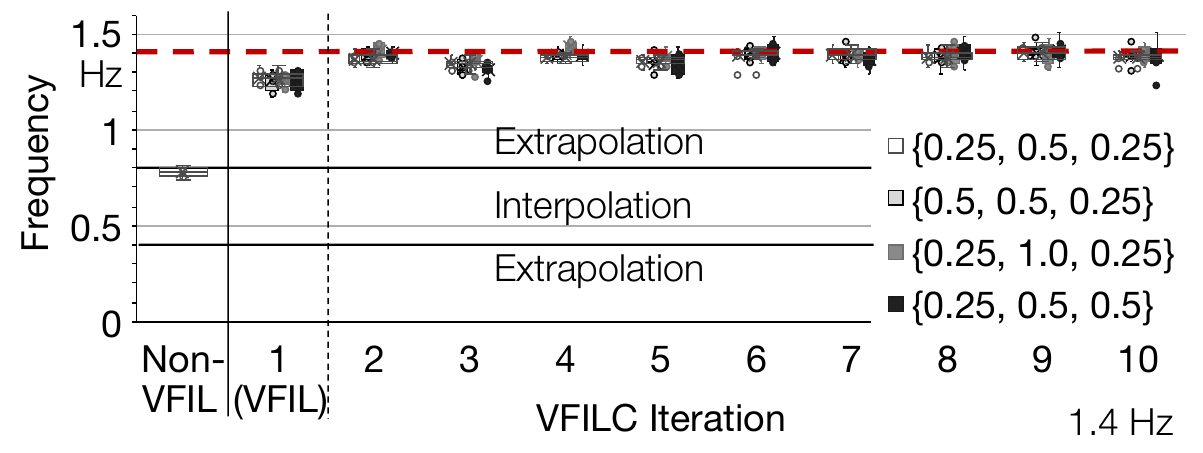}}
                    \label{wp14freq2}
                }
                \caption{Real motion frequencies in the wiping task, with four different ILC gains. Red dashed lines indicate the command values. The plot (a) indicates the traits of each gain values with different variances, whereas the plot (b) shows the overall similarity approaching 1.4~Hz.}
                \label{fig:wpfreq2}
            \end{figure}

\begin{figure}[tbp]
                \centering
                \subfloat[0.6 Hz]{%
                    
                    \resizebox*{8.8cm}{!}{\includegraphics[scale=0.6, bb = 0 0 577 217]{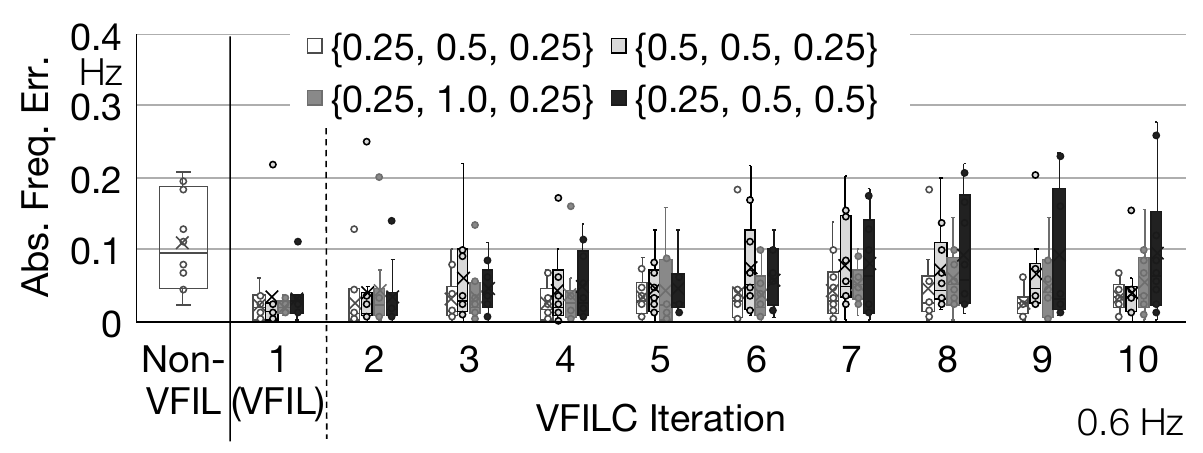}}
                    \label{wp06abs2}
                }
                \vspace{0.5pt}
                \centering
                \subfloat[1.4 Hz]{%
                    
                    \resizebox*{8.8cm}{!}{\includegraphics[scale=0.6, bb = 0 0 577 217]{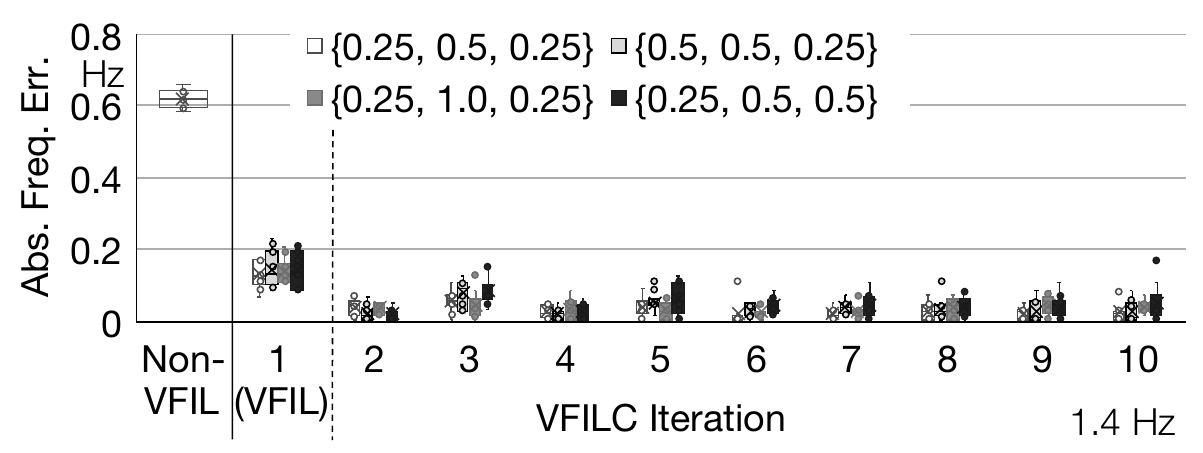}}
                    \label{wp14abs2}
                }
                \caption{Absolute frequency errors in the wiping task, with four different ILC gains. While the plot (b) shows an overall convergence, the plot (a) shows the larger variances caused by overshoot and oscillation in the large-$K_i$ and the large-$K_p$ setting, respectively.}
                \label{fig:wpabs2}
            \end{figure}

            Figs.~\ref{fig:wpfreq2} and~\ref{fig:wpabs2} compare the real frequencies and absolute frequency errors in the wiping task between VFILC with gains $\{ K_p, K_i, K_d \} = \{ 0.25, 0.5, 0.25 \} $ and the non-VFIL setting.
            In the interpolated 0.6~Hz setting, the average frequency errors were continuously below 0.05~Hz from the first iteration, indicating the successful feedforward compensation and the stable feedback.
            As for the basic gain set $\{ K_p, K_i, K_d \} = \{ 0.25, 0.5, 0.25 \} $, the average absolute frequency error in 1.4~Hz setting, shown in Fig.~\ref{fig:wpabs2}~\subref{wp14abs2}, was reduced from 0.13~Hz in the first iteration to 0.025~Hz in the tenth iteration, achieving a remarkable 81\% decrease. 
            Compared to non-VFIL trials limited in interpolations, the error decreased by 96\%.

            In comparisons with other gain sets, the large-$K_p$ setting experienced overshooting in early iterations, whereas the large-$K_i$ setting had larger variances in late iterations.
            In the large-$K_d$ setting, especially in Fig.~\ref{fig:wpfreq2}~\subref{wp06freq2}, oscillations were observed. This is mainly due to the combination of a lack of a low-pass filter and delays in calculation of $K_d$-based inputs.

            The overall success rate of VFILC setting was 90\%. A total of 89 iterations reached the speed limit before reaching the 400 successful iterations in the 1.4~Hz setting. No other types of failures were observed, including the 0.6~Hz setting with another 400 successful iterations. 
            No failures were observed in non-VFIL setting, mainly due to its slow speed.


        \subsubsection{Mixing Task}
\begin{figure}[tbp]
                \centering
                \subfloat[0.5 Hz]{%
                    
                    \resizebox*{8.8cm}{!}{\includegraphics[scale=0.6, bb = 0 0 577 217]{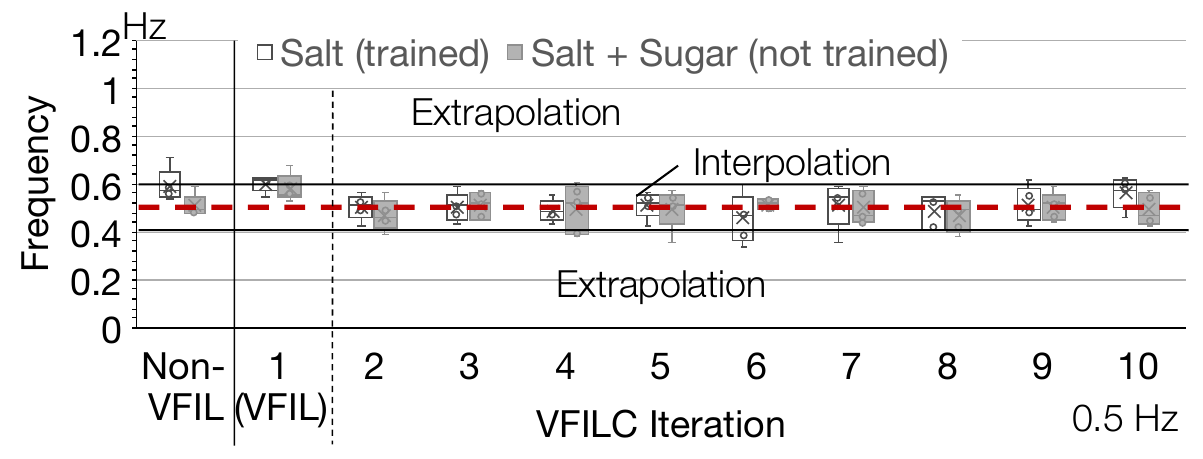}}
                    \label{mx05freq}
                }
                \vspace{0.5pt}
                \centering
                \subfloat[1.0 Hz]{%
                    
                    \resizebox*{8.8cm}{!}{\includegraphics[scale=0.6, bb = 0 0 577 217]{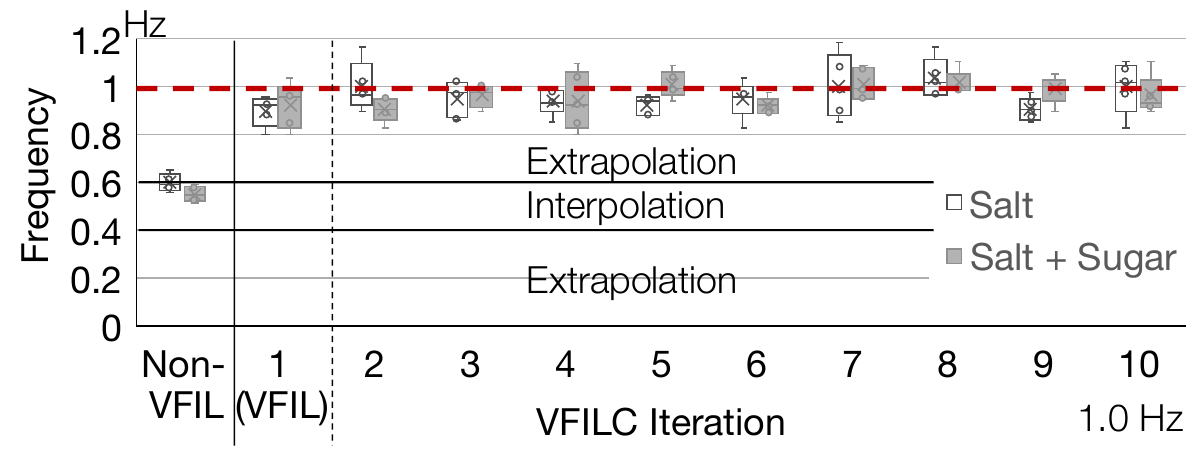}}
                    \label{mx10freq}
                }
                \caption{Real motion frequencies in the mixing task. Red dashed lines indicate command values. The plot (a) shows the fluctuation and the converging trend of motion frequencies over the iterations, while the plot (b) indicates the overall stability in the out-of-distribution, high-frequency 1.0~Hz setting.}
                \label{fig:mxfreq}
            \end{figure}

\begin{figure}[tbp]
                \centering
                \subfloat[0.5 Hz]{%
                    
                    \resizebox*{8.8cm}{!}{\includegraphics[scale=0.6, bb = 0 0 577 217]{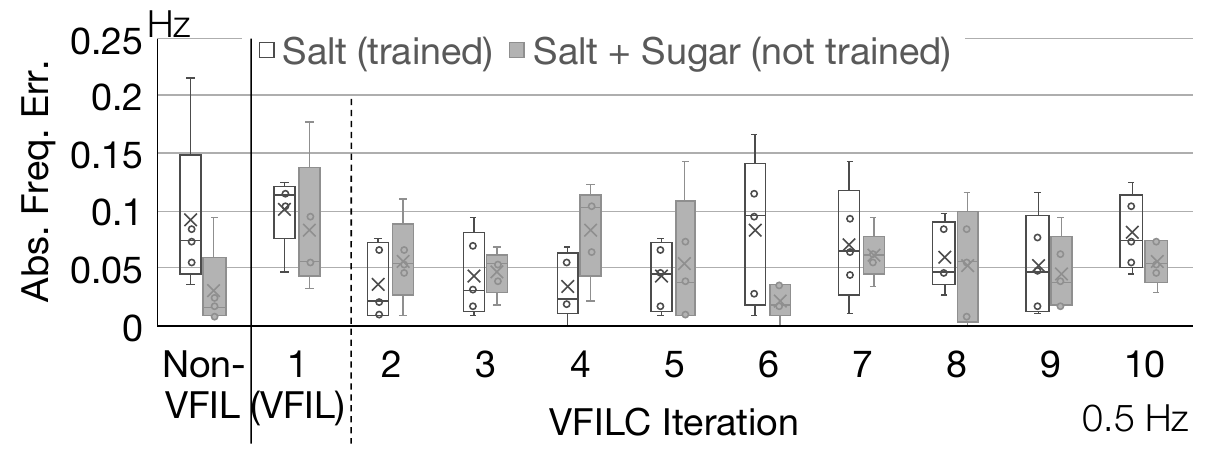}}
                    \label{mx05abs}
                }
                \vspace{0.5pt}
                \centering
                \subfloat[1.0 Hz]{%
                    
                    \resizebox*{8.8cm}{!}{\includegraphics[scale=0.6, bb = 0 0 577 217]{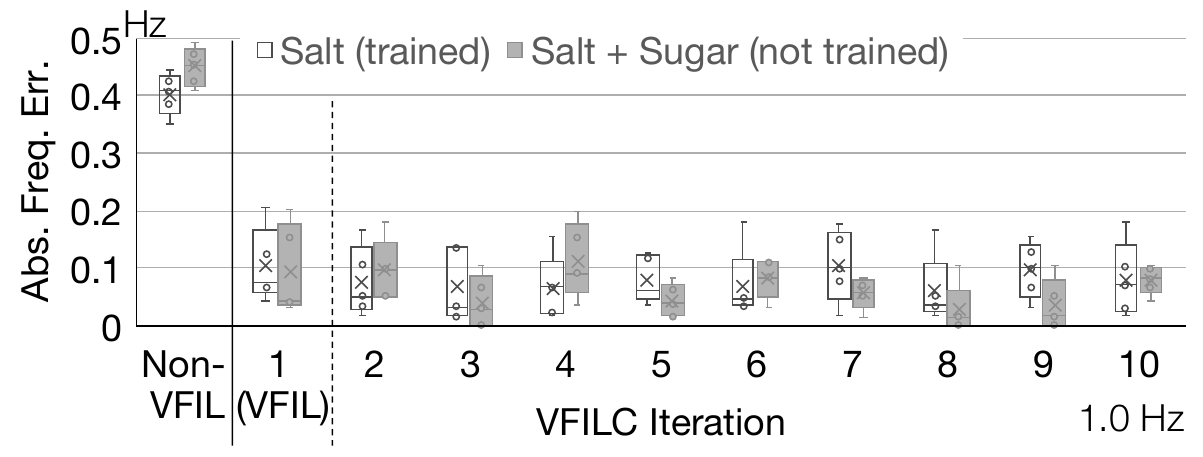}}
                    \label{mx10abs}
                }
                \caption{Absolute frequency errors in the mixing task. The plot (a) shows the adjustment of frequency errors over the iterations, while the plot (b) shows the overall stability.}
                \label{fig:mxabs}
            \end{figure}

            Figs.~\ref{fig:mxfreq} and~\ref{fig:mxabs} depict the real frequencies and absolute frequency errors in the mixing task. 
            The VFIL and VFILC iterations' similarity in both the trained salt-only and unseen salt-and-sugar settings depicts the high adaptability to environmental changes.
            In the 0.5~Hz setting, the initial motion frequency was approximately 0.6~Hz rather than 0.5~Hz command. The feedback reduced the error, especially in the second iteration. 
            Larger absolute errors in the following iterations, as shown in Fig.~\ref{fig:mxabs}~\subref{mx05abs}, were due to fluctuating motion frequencies. 
            Despite this fluctuation, the overall average absolute frequency error of 0.5~Hz setting decreased by 27\% throughout ILC, from 0.092~Hz to 0.067~Hz.
            Unlike other tasks, changes in frequency errors were smaller in the extrapolated 1.0~Hz setting than in the trained 0.5~Hz setting, likely due to powder's complicated traits of friction.
            As shown in Fig.~\ref{fig:mxabs}~\subref{mx10abs}, in the 1.0~Hz setting, the feedforward enabled accurate speed extrapolations from the first iteration, keeping the average absolute frequency error below 0.1~Hz, compared to 0.43~Hz in the non-VFIL setting.
            
\begin{figure}[tbp]
                \centering
                \subfloat[Typical spilling]{%
                    \resizebox*{!}{4cm}{\includegraphics[scale=0.6, bb = 0 0 559 905]{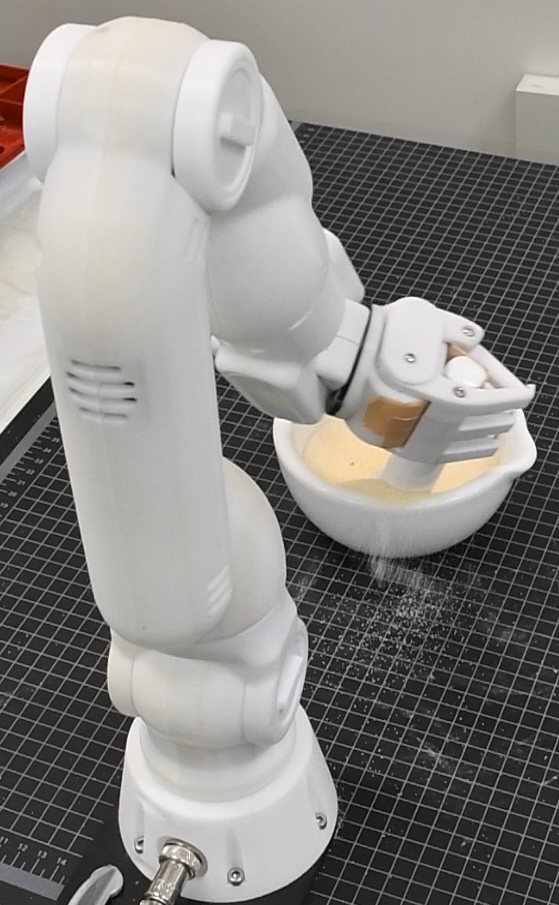}}
                    \label{typical}
                }
                \hspace{5pt}
                \subfloat[Knocking down]{%
                    \resizebox*{!}{4cm}{\includegraphics[scale=0.6, bb = 0 0 696 963]{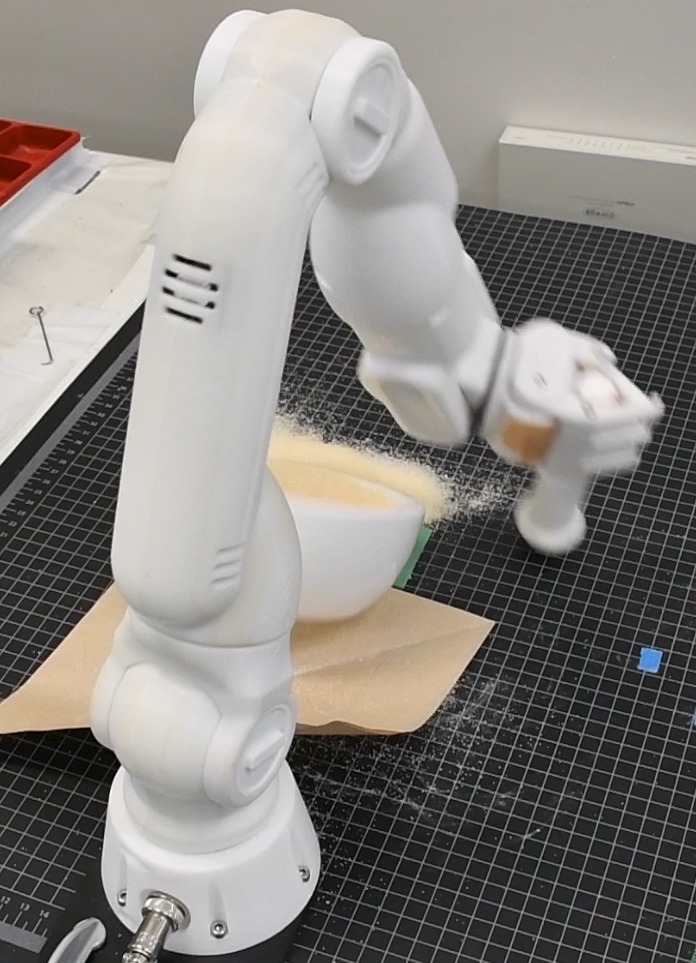}}
                    \label{worst}
                }
                \caption{Spilling observed in the mixing task.}
                \label{fig:spilling}
            \end{figure}

            After the addition of sugar, spillage was observed, as shown in Fig.~\ref{fig:spilling}, in both non-VFIL and VFILC settings. 
            In non-VFIL trials, the maximum spillage per trial was 5.4 g.
            In VFILC trials, one failed iteration in the 1.0~Hz setting caused the maximum cumulative spillage per trial to reach 29.7~g by knocking down the mortar. Except for this outlier, the maximum spillage per trial caused by mixing was 4.4~g across the 10 iterations.

            The success rate of VFILC iterations was 81\%, whereas non-VFIL iterations had 67\%. the VFILC setting experienced 47 failed iterations before reaching 200 successful ones, whereas the non-VFIL setting experienced five failures before 10 successful iterations.

                                    
        \subsubsection{Shaking Task}
\begin{figure}[tbp]
                \centering
                \subfloat[1.0 Hz]{%
                    
                    \resizebox*{8.8cm}{!}{\includegraphics[scale=0.6, bb = 0 0 577 217]{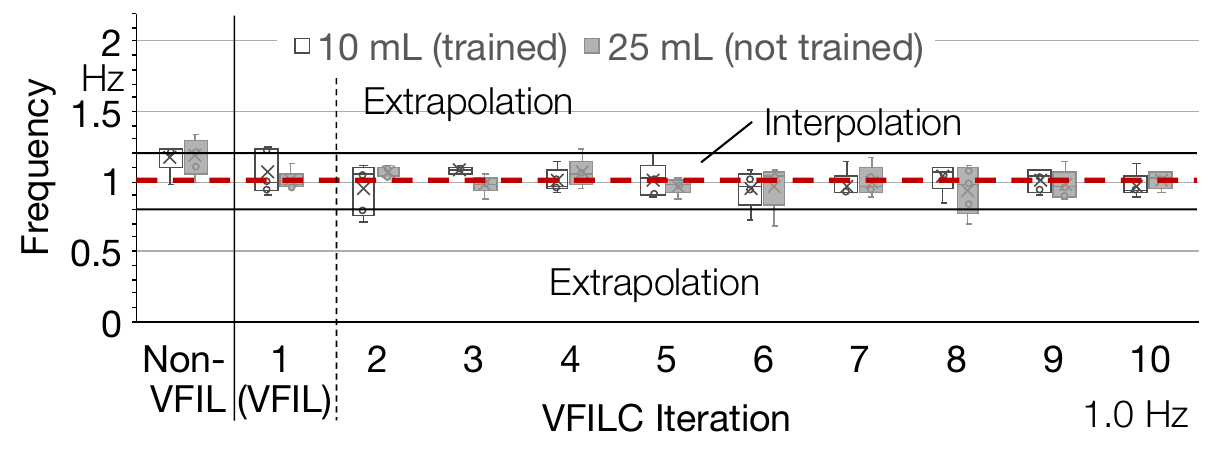}}
                    \label{sh10freq}
                }
                \vspace{0.5pt}
                \centering
                \subfloat[2.0 Hz]{%
                    
                    \resizebox*{8.8cm}{!}{\includegraphics[scale=0.6, bb = 0 0 577 217]{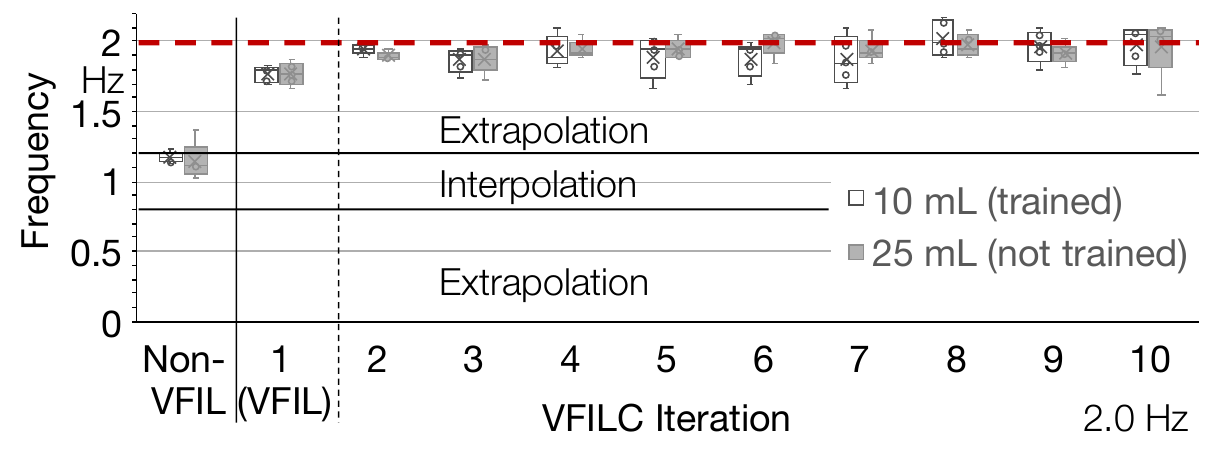}}
                    \label{sh20freq}
                }
                \caption{Real motion frequencies in the shaking task. Red dashed lines indicate command values. The plot (a) shows the overall stability along the 1.0~Hz command from the first iteration, and the plot (b) indicates the upward trend towards the 2.0~Hz command with the repeated feedback along with iterations.}
                \label{fig:shfreq}
            \end{figure}

\begin{figure}[tbp]
                \centering
                \subfloat[1.0 Hz]{%
                    
                    \resizebox*{8.8cm}{!}{\includegraphics[scale=0.6, bb = 0 0 577 217]{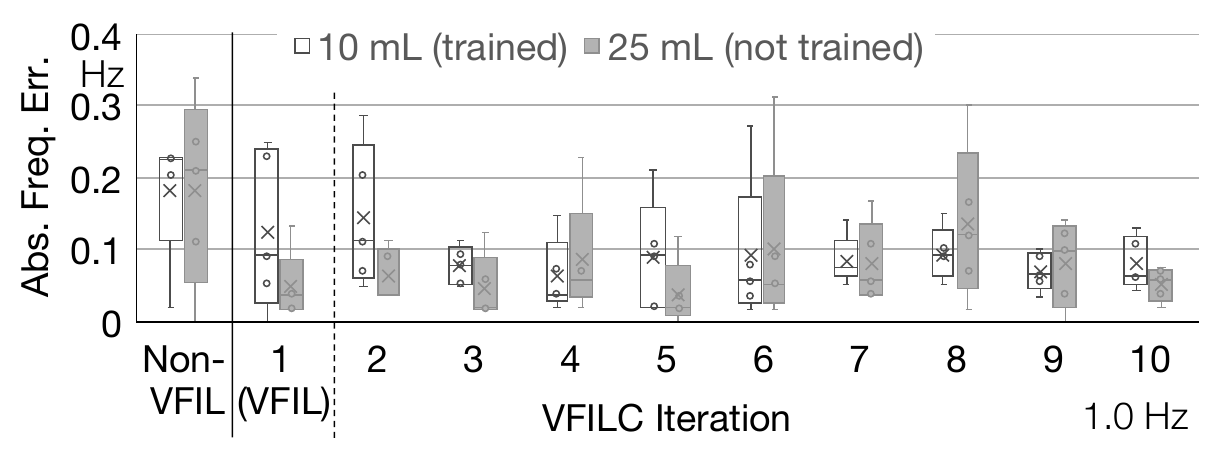}}
                    \label{sh10abs}
                }
                \vspace{0.5pt}
                \centering
                \subfloat[2.0 Hz]{%
                    
                    \resizebox*{8.8cm}{!}{\includegraphics[scale=0.6, bb = 0 0 577 217]{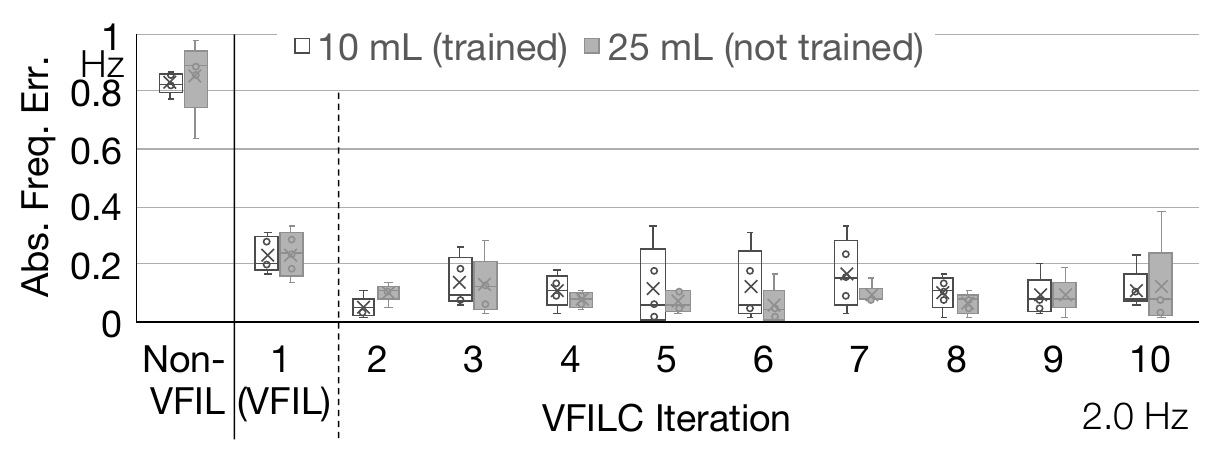}}
                    \label{sh20abs}
                }
                \caption{Absolute frequency errors in the shaking task. The plot (b) especially shows the decreasing errors over the iterations.}
                \label{fig:shabs}
            \end{figure}

            Figs.~\ref{fig:shfreq} and~\ref{fig:shabs} depict the real frequencies and absolute frequency errors in the shaking task.
            Careful and slow grasping and pulling procedures made shaking periods shorter, contributing to a smaller sample size and larger frequency errors especially at the 1.0~Hz setting.
            The overall similarity between the trained 10~mL and unseen 25~mL settings in the real frequencies depicts the proposed method's robustness against environmental changes.

            In the trained 1.0~Hz setting, as shown in Fig.~\ref{fig:shfreq}~\subref{sh10freq}, the shaking frequency fluctuated over iterations, suggesting room for gain tuning and refining the datasets.
            As shown in Fig.~\ref{fig:shfreq}~\subref{sh20freq}, the average shaking frequency at the untrained 2.0~Hz setting moved from 1.77~Hz in the first iteration to 1.97~Hz in the final iteration, decreasing frequency error over iterations due to feedback.
            Fig.~\ref{fig:shabs}~\subref{sh20abs} illustrates the average absolute frequency error over the iterations, a 50\% decrease from 0.23 Hz at the first iteration to 0.11 Hz at the tenth iteration, and an 86\% decrease from non-VFIL.
            
            The success rate of VFILC iterations was 98\%, whereas non-VFIL iterations had a 77\% success rate. Failures in grasping were observed in five iterations with VFILC trials and three with non-VFIL iterations.

\begin{figure}[tbp]
                \centering
                \subfloat[Angled pulling]{%
                    \resizebox*{!}{4cm}{\includegraphics[scale=0.6, bb = 0 0 658 959]{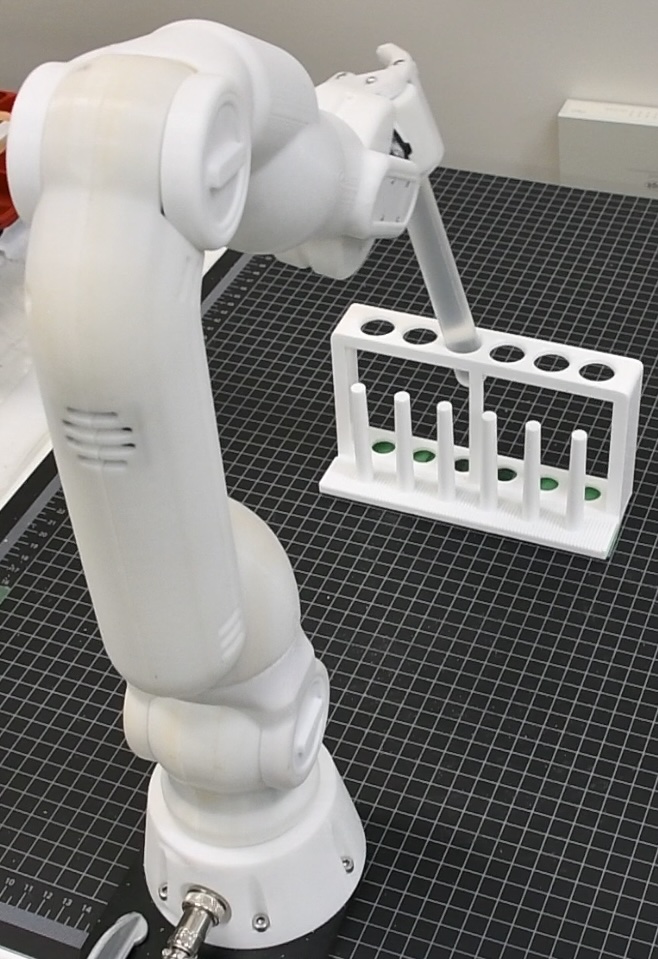}}
                    \label{anglepulling}
                }
                \hspace{5pt}
                \subfloat[Removing the rack]{%
                    \resizebox*{!}{4cm}{\includegraphics[scale=0.6, bb = 0 0 722 978]{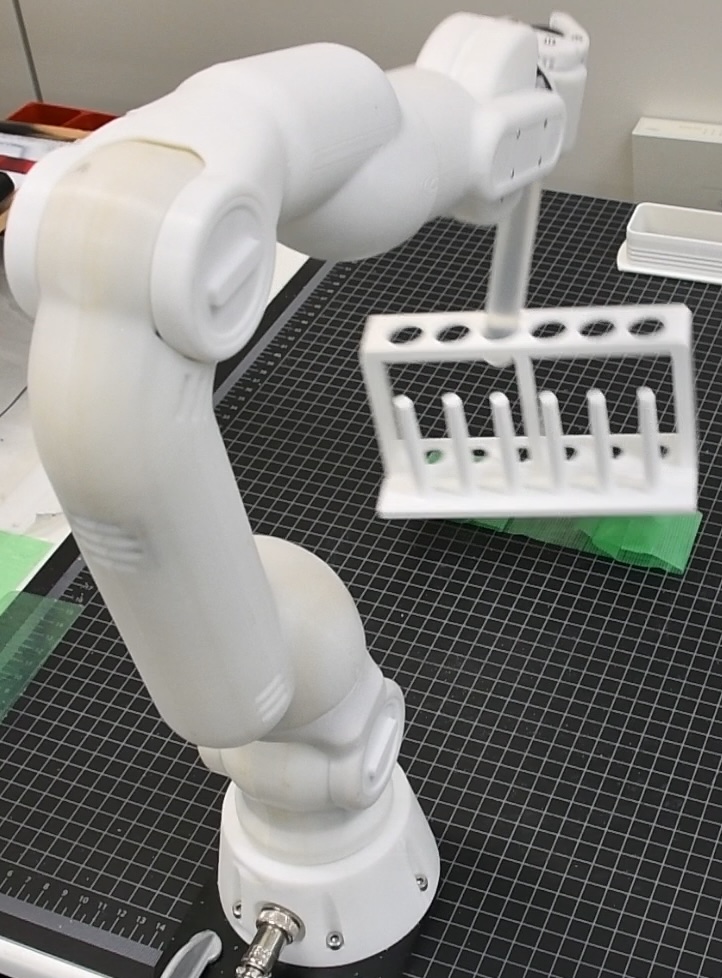}}
                    \label{removing}
                }
                \caption{Mixed behaviors between pulling and shaking.}
                \label{fig:conflation}
            \end{figure}

            Pulling at an angle, which was not in the training data, was frequently observed regardless of water volumes or shaking frequencies, as shown in Fig.~\ref{fig:conflation}. This resulted in removing the test tube rack in two iterations, while both did not interfere with the process and stability of ILC. 
            This suggests a conflation of behavior between the trained, vertical pulling phase and the following shaking phase.


    \subsection{Discussions}
        Overall, the proposed method achieved comparable accuracy between interpolated and extrapolated motion frequencies across all three tasks.
        Especially, decreased frequency errors at high-frequency settings of wiping and shaking tasks, 81\% and 50\% each, suggest that the proposed method succeeded in adjusting motion frequencies, which are particularly affected by significant changes due to nonlinearity.
        Given that the PID gains of ILC were not individually tuned to each task, more improvements can be expected in the mixing and shaking tasks with adequate gain tuning, in addition to more precise training datasets and self-supervised learning~\cite{CRANEX7params}.
        Also, the feasibility and stability of VFIL against both temporal and environmental changes were verified with a total of 1,200 successful iterations, of which 400 were previously unverified tasks.

        Below, we describe the limitations of VFILC.

        First, the efficiency of this method can decrease in low-frequency settings, because the base method VFIL is more prone to failures at low frequencies due to the increased sampling time restricting adaptability.
        Potential methods to mitigate this are switching NN models for multiple sampling frequencies or interpolations between samples, as discussed in the original VFIL paper~\cite{VFIL}.

        Second, because this method is largely based on fast-forwarding and not simulation, the risk of unexpected motion persists, such as exceeding speed limits, spillage, or grasping failures, especially with combinations of multiple untrained conditions. 
        Failure predictions with visual data and simulation-based methods, in addition to human interventions~\cite{chgdagger}, can resolve this issue. 
        
        Third, because this method adjusts the motion frequency at each iteration, delays in each step of motion are not adjusted separately from other steps, even in repetitive components of each task, such as one round trip of wiping or pulling of the test tube.
        Not only does this hinder precise motion at the frequency command, this can also cause conflation between components with varying durations, as observed between pulling and shaking in the shaking task. 
        This lengthiness of iterations also limits the effect of derivative control, which requires low-pass filters with lower cutoff frequencies to prevent oscillations. 
        A combination of fine-grained motion segmentation, per-primitive normalization and feedback will improve the overall quality of generated motion. 
        This can also contribute to the adaptations of VFILC in non-periodic, multi-step, yet repetitive tasks.

\section{Conclusion}
    This paper proposes VFILC, a method that uses iterative learning control to adjust the motion speed in variable-frequency imitation learning~(VFIL), adjusting both the frequency label and the sampling frequency of the NN model.
    This implementation of feedback along with feedforward has improved temporal accuracy of extrapolated high-frequency contact-rich tasks, whose nonlinearity significantly affected the motion frequency under VFIL, while preserving VFIL's accuracy and adaptability to situations with smaller errors.
    The experimental results showed that the proposed method enabled an equally accurate extrapolation of motion speeds compared to interpolation in all three tasks, with substantial 81\% and 50\% decreases of frequency errors in the wiping and shaking tasks respectively at extrapolated high-speed settings, as well as a 27\% reduction in the mixing task at an interpolated speed.
    Our future work will include adaptations to low-frequency ranges, failure predictions and more fine-grained feedback, to further improve the accuracy and quality of variable-speed motions, including non-periodic ones.

\bibliographystyle{IEEEtran}

\end{document}